\documentclass[10pt,twocolumn,letterpaper]{article}

\usepackage[pagenumbers]{cvpr} % To force page numbers, e.g. for an arXiv version

\usepackage{multirow} % Required for multirows

\definecolor{cvprblue}{rgb}{0.21,0.49,0.74}
\usepackage[pagebackref,breaklinks,colorlinks,allcolors=cvprblue]{hyperref}

\def\confName{CVPR}
\def\confYear{2027}

\title{GauBlend: Dynamic Gaussian Blending Bases via One-Step Diffusion \\for One-Shot Animatable Head Avatars}
\author{Shuling Zhao$^{1,2}$, \ \ Andrew Hung$^1$, \ \ Boyao Zhou$^2$, \ \ Dan Xu$^{1}$\\
\vspace{-13pt}
\and
$^1$The Hong Kong University of Science and Technology, \quad $^2$Ant Group\\
}

\begin{document}
% \maketitle
\twocolumn[{%
\renewcommand\twocolumn[1][]{#1}%
\maketitle

\begin{center}
    \centering
    \captionsetup{type=figure}
  \includegraphics[width=0.95\textwidth]{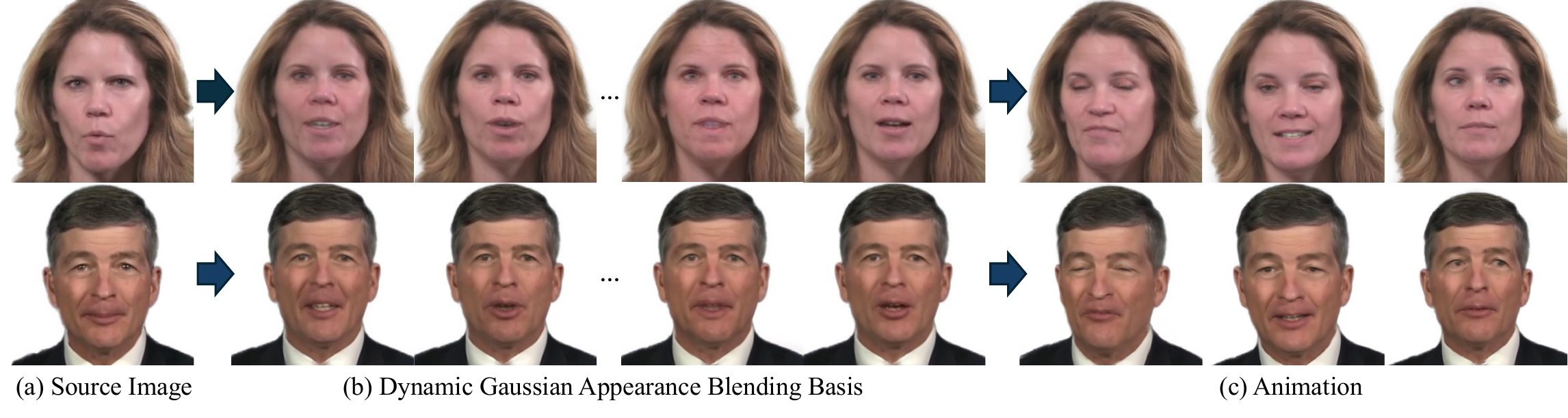}
  \vspace{-10pt}
  \caption{Overview of GauBlend. (a) From a single source image, GauBlend generates a canonical Gaussian avatar with a dynamic appearance blending basis via one-step diffusion. (b) Varying only the blending weights produces diverse expression-related appearances (see the lower face and forehead). (c) The basis is efficient blended with driving expressions, enabling natural animation at 114 FPS.
%  during animation, producing natural, expression-dependent details at 114 FPS.
  }
  \label{fig:teaser}
\end{center}
}]

\begin{abstract}
One-shot animatable head avatar reconstruction aims to reconstruct a 3D head from a single image and animate it under novel motions. Although recent feed-forward methods have advanced the field with large regression-based transformers, it remains challenging to efficiently model detailed, expression-dependent appearance from a single input image. Consequently, they often produce avatars with stiff expressions and slow animation. In this work, we propose GauBlend, a one-step diffusion model that generates a canonical Gaussian avatar with a dynamic appearance blending basis, enabling efficient dynamic appearance modeling during animation. To address the ill-posed nature of one-shot reconstruction, we integrate a 3D canonical avatar representation into a 2D diffusion pipeline, directly generating, animating, and rendering it at each timestep during training. This enables effective 2D supervision while preserving 3D consistency. Furthermore, we design a dynamic Gaussian blending basis that explicitly contains the dynamic appearance variations of the input subject to enhance the canonical avatar. The basis can be efficiently blended with a driving expression code to model expression-dependent details, resulting in more natural and faster avatar animation. We further improve reconstruction fidelity through pixel-aligned Gaussian regularization and a two-stage training strategy. At inference, GauBlend generates the canonical Gaussian avatar in a single step. Extensive experiments show that GauBlend achieves state-of-the-art performance in one-shot feed-forward animatable head reconstruction and supports animation at 114 FPS. \footnote{\href{https://shaelynz.github.io/gaublend_project/}{https://shaelynz.github.io/gaublend\_project/}}
\end{abstract}    
\vspace{-11pt}
\section{Introduction}
\label{sec:intro}
\vspace{-3pt}

Reconstructing animatable 3D head avatars from a single image is a fundamental challenge in digital avatar creation, with promising applications in remote communication, gaming, and virtual reality. Recent approaches predominantly utilize 3D Gaussian Splatting (3DGS)~\cite{kerbl20233d} as their underlying representation because of its high visual quality and fast rendering speed.

A line of work~\cite{taubner2025cap4d, taubner2025mvp4d, yin2025facecraft4d} uses a two-stage pipeline that first generates multi-view images with diverse expressions via 2D diffusion models and then reconstructs a Gaussian head avatar from them. 
%However, separating image generation from avatar reconstruction can cause 3D inconsistencies that degrade reconstruction quality. Moreover, iterative diffusion sampling and 
%the iterative diffusion sampling and subsequent identity-specific reconstruction are computationally expensive, often requiring several hours at inference time. To improve inference efficiency, 
However, this identity-specific two-stage reconstruction process is time-consuming and may introduce 3D inconsistencies. To avoid these problems, existing feed-forward methods~\cite{he2025lam, wu2026uika, liu2026avatarpointillist, zhao2026one} typically reconstruct Gaussian avatars directly by predicting canonical Gaussian attributes from the input image. They then animate the avatars by deforming Gaussian positions according to a template mesh~\cite{li2017learning}. Despite their fast animation speeds, these methods do not model expression-dependent appearance changes, such as wrinkles and shadows, resulting in stiff and unnatural expressions during animation. 
To enable more natural motion, some methods~\cite{kirschstein2026flexavatar, peng2026flexavatar} predict neutral avatar features from the input image and fuse them with expression signals using attention- or UNet-based Gaussian decoders to generate expression-conditioned Gaussian attributes. However, neutral features cannot adequately preserve the dynamic appearance information from the input image. Moreover, the computational overhead introduced by these decoder architectures significantly reduces animation speed.
%However, they ignore the dynamic appearance information from the input image for animation, and the attention or UNet designs slow down the animation speed.
%To avoid this problem, some methods~\cite{kirschstein2025avat3r, kirschstein2026flexavatar} inject the expression code into a Gaussian decoder via cross-attention to generate expression-conditioned Gaussian attributes. However, the cross-attention mechanism prevents real-time animation. FlexAvatar~\cite{peng2026flexavatar} predicts motion-related delta Gaussian maps with a UNet, but it ignores the deformation information from the input image that may help model expression-related dynamic details.
Existing feed-forward methods also predominantly rely on regression-based paradigms, which often produce less expressive results.

%To address these issues, %inspired by the success of one-stage generative models in 3D reconstruction~\cite{cai2025baking} and blendshape models in identity-specific avatar reconstruction~\cite{ma20243d, dhamo2024headgas, li2025rgbavatar}, 
% we draw inspiration from the concept of blendshapes in 3D head avatars~\cite{blanz1999morphable, dhamo2024headgas} and extend it to generalizable dynamic appearance modeling. We propose a one-step, generation-based framework for one-shot animatable head avatar reconstruction. Specifically,  our framework uses a 2D image denoiser to generate images of the input person with novel expressions and poses from a single input image in a single forward pass. This is realized by predicting a canonical Gaussian avatar representation with an expression-related Gaussian blending basis, which is then animated under target motions and poses. Directly generating a Gaussian representation ensures 3D consistency, while denoising images with novel motions and viewpoints brings additional dynamic appearance cues for predicting a more expressive Gaussian blending basis. During animation, we efficiently obtain dynamic Gaussian colors and opacities by linearly blending the generated basis features with a compressed expression code, enabling both detailed dynamic appearance modeling and real-time animation. 
% we draw inspiration from the concept of blendshapes in 3D head avatars~\cite{blanz1999morphable, dhamo2024headgas} and extend it to generalizable dynamic appearance modeling. 
To achieve natural dynamic appearance modeling with fast animation, we propose GauBlend, a one-step diffusion model that directly generates a canonical Gaussian avatar with a dynamic appearance blending basis for one-shot animatable head avatar reconstruction. Specifically, to address the ill-posed nature of one-shot reconstruction and produce expressive canonical avatars with dynamic appearance, we integrate the canonical avatar representation into a 2D diffusion pipeline. During training, the diffusion model directly generates the canonical avatar at each timestep, then animates and renders it to denoise images of the input subject with novel expressions and poses. This enables supervision from 2D video frames while enforcing 3D consistency. Moreover, we design a dynamic Gaussian blending basis that explicitly models dynamic appearance variations of the input subject to enhance the canonical avatar for different expressions (see Fig.~\ref{fig:teaser}b). The basis can be efficiently linearly blended with a driving expression code and decoded into Gaussian colors and opacities using a lightweight MLP. This enables detailed, expression-dependent appearance modeling, resulting in fast and natural avatar animation. To further improve reconstruction fidelity, we introduce pixel-aligned Gaussian regularization and a two-stage training strategy. At inference time, GauBlend efficiently generates the canonical Gaussian avatar in a single step. 

% we train our model to denoise images of the input person with novel expressions and poses from a single input image at every timestep. This is realized by animating the canonical avatar under target motions and poses. Directly generating a Gaussian representation ensures 3D consistency, while denoising images with novel motions and viewpoints brings additional dynamic appearance cues for predicting a more expressive Gaussian blending basis. Our model supports fast one-step inference. During animation, we efficiently obtain dynamic Gaussian colors and opacities by linearly blending the generated basis features with a compressed expression code, enabling both detailed dynamic appearance modeling and real-time animation. 

We conduct extensive experiments to evaluate the effectiveness of GauBlend. The results show that our method produces expressive Gaussian blending bases for expression-dependent appearance modeling and achieves state-of-the-art performance in one-shot feed-forward animatable avatar reconstruction. We summarize our contributions as follows:

\begin{itemize}[leftmargin=*]
    \item %We propose a novel one-step generative framework for reconstructing animatable Gaussian head avatars from a single image. Our framework uses an image denoiser to generate a canonical 3D Gaussian avatar representation, and animates it to produce denoised images of the input subject with novel poses and expressions. Denoising these images provides additional appearance and motion cues that facilitate the generation of the canonical Gaussian representation.
    % We propose a novel one-step diffusion framework for reconstructing animatable Gaussian head avatars from a single image. Our framework uses an image denoiser to generate a canonical 3D Gaussian avatar representation, and animates it to produce denoised images of the input subject with novel poses and expressions. Denoising these images provides additional appearance and motion cues that facilitate the generation of the canonical Gaussian representation.
    We propose GauBlend, a novel one-step diffusion model for reconstructing animatable Gaussian head avatars from a single image. 
    By integrating a canonical Gaussian representation into a 2D diffusion pipeline, GauBlend directly generates the canonical avatar at each timestep during training, enabling effective 2D supervision, 3D consistency, and efficient one-step inference.
    %Our model directly generates a canonical 3D Gaussian avatar representation. We train our model to denoise 2D images of the input subject with novel poses and expressions by animating the canonical avatar. Denoising these images provides additional appearance and motion cues that facilitate the generation of the canonical Gaussian representation. Our model supports one-step generation at inference.
    %     generate images of the input subject with novel poses and expressions, providing additional appearance and motion cues for the simultaneous generation of a 3D Gaussian avatar representation.
    % our framework employs an image denoiser to produce a canonical Gaussian avatar representation with an expression-related blending basis and animates it to generate images of the input subject with novel expressions and poses.
    \item Instead of using static Gaussian colors and opacities to model avatar appearance, we incorporate a dynamic Gaussian blending basis into the canonical representation for efficient, expression-dependent appearance modeling. The basis is blended with a driving expression code and decoded into expression-dependent Gaussian colors and opacities using a lightweight MLP, enabling more natural animation at higher speeds.
    \item Extensive experiments on large-scale datasets demonstrate that GauBlend produces expressive Gaussian blending bases for dynamic appearance modeling, achieves state-of-the-art one-shot feed-forward 3D avatar animation quality, and supports efficient animation at 114 FPS.

\end{itemize}
% [training, + one-sentence inference, fast, fps]

%To further improve 3D consistency at large views, we utilize a two-stage training pipeline. Specifically, we use large

%novel expressions of the person in the input image at different poses, and 

%generate an expression-related Gaussian blending basis from a single image with a one-step image denoiser for animatable avatar reconstruction. 
% During animation, we linearly blend the learned basis features with a compressed expression code to realize both expression-related appearance changes and real-time animation simultaneously. Specifically, we use a one-step 
% \vspace{-6pt}
\section{Related Work}
\label{sec:related_work}
% \vspace{-6pt}
% \vspace{-5pt}

\begin{figure*}[tb]
  \centering
  \includegraphics[width=0.91\linewidth]{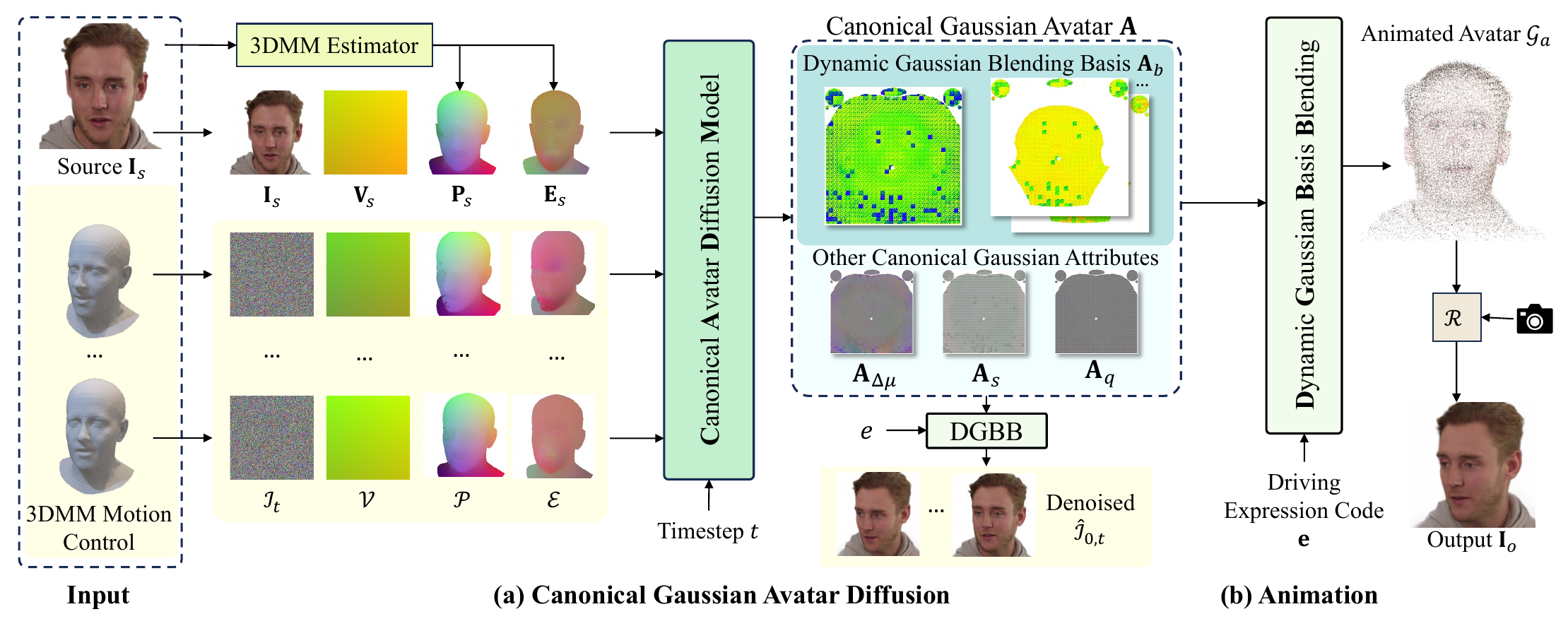
  }
  %{figures/overview_full_shadow.pdf}
  \vspace{-9pt}
  \caption{Framework of GauBlend. Given a source image, (a) \textbf{Canonical Avatar Diffusion Model (CADM)} uses it as the image condition, along with camera-view, head-pose, and expression maps for the source image and 3DMM motions for the noisy images as additional conditions. It directly generates a UV-space canonical Gaussian avatar with a dynamic Gaussian blending basis at each timestep, which is then animated and rendered to denoise the images during training. (b) During animation, \textbf{Dynamic Gaussian Basis Blending (DGBB)} uses a driving expression code to transform the canonical representation into an animated avatar with expression-dependent appearance, which can then be rendered into an output image under a given camera pose.
  }
  %{Overview of the framework. (a) Canonical Gaussian Avatar Diffusion takes a source image as the image condition, together with the corresponding camera-view, head-pose, and expression maps for the source and noisy images as additional conditions. It directly generates a UV-space canonical Gaussian representation with a blending basis. During training, it also produces pixel-aligned Gaussians for regularization, thereby improving the reconstruction fidelity of the canonical representation. (b) During animation, the Gaussian Basis Blending module transforms the canonical representation into an animated avatar with expression-dependent dynamic appearances, which can then be rendered into an output image under a given camera pose.
  % }
  \label{fig:overview}
  \vspace{-10pt}
\end{figure*}

% \noindent\textbf{3D Animatable Head Avatars from Sparse Views.}
\noindent\textbf{One-Shot 3D Animatable Head Avatars.}
Existing methods for %sparse-view 
one-shot 3D animatable head avatar reconstruction can be broadly divided into optimization-based and feed-forward approaches. Optimization-based methods~\cite{taubner2025cap4d, taubner2025mvp4d, yin2025facecraft4d, zhou2025zero} use powerful multi-view or video generative models~\cite{rombach2022high, guo2024liveportrait, ma2024follow} to synthesize images of the input subject with diverse expressions and viewpoints, then optimize an avatar from them. Although effective, the identity-specific two-stage reconstruction is time-consuming and may introduce 3D inconsistencies. In contrast, feed-forward methods~\cite{khakhulin2022realistic, li2023generalizable, ye2024realdportrait, deng2024portrait4d, deng2024portrait4d_eccv} reconstruct avatars in a single forward pass and generalize to unseen identities through training on large-scale image or video datasets~\cite{zhu2022celebv, chung18b_interspeech, karras2019style}. Recent approaches~\cite{chu2024generalizable, he2025lam, kirschstein2026flexavatar, peng2026flexavatar, wu2026uika, ren2026omg, liu2026avatarpointillist, zhao2026one, yao2026ffavatar} further improve efficiency by adopting 3DGS~\cite{kerbl20233d}. They typically predict a canonical Gaussian avatar from the input image and deform it using a FLAME mesh~\cite{li2017learning}. However, they often fail to model expression-dependent appearance, resulting in stiff animation. Some methods~\cite{chu2024generalizable, ren2026omg} employ 2D neural renderers to enhance motion-related details, but this may compromise 3D consistency. Others~\cite{peng2026flexavatar, yao2026ffavatar} use refinement networks to predict motion-related Gaussian offsets. FlexAvatar~\cite{kirschstein2026flexavatar} directly injects expression codes into a Gaussian decoder via cross-attention to produce avatars with target motions. However, these methods introduce considerable computational overhead during animation, substantially reducing their speed. In contrast, our diffusion transformer directly generates dynamic Gaussian blending bases. A lightweight MLP then linearly combines the blending basis with an expression code to produce expression-dependent Gaussian colors and opacities, enabling efficient dynamic appearance modeling.
%suffer from a substantial reduction in animation speed because of the considerable computational overhead required to apply new motions. In contrast, we make full use of the diffusion transformer architecture to directly generate expression-related Gaussian blending bases as well as other Gaussian attributes in the canonical space. This allows us to use a small MLP to linearly combine the blending bases with a compressed motion code and produce expression-related Gaussian color and opacity values, effectively modeling dynamic appearances with higher efficiency.

\noindent\textbf{Blendshape-based Face Modeling and Animation.}
Blendshapes are widely used in facial modeling for their efficiency and expressiveness. The seminal 3D Morphable Model (3DMM)~\cite{blanz1999morphable} represents 3D faces as linear combinations of face prototypes, enabling low-dimensional control of shape and texture. Subsequent works improve expressiveness through multilinear representations~\cite{vlasic2006face, cao2013facewarehouse}, articulated poses~\cite{li2017learning}, more diverse and accurate scans~\cite{paysan20093d, cao2013facewarehouse, li2017learning}, and complete head models~\cite{ploumpis2020towards}. Other approaches~\cite{yenamandra2021i3dmm, hong2022headnerf, giebenhain2023learning} extend parametric models to implicit representations to overcome the limited accuracy of meshes. However, these methods typically require identity-code fitting before animation, and their reconstruction quality is constrained by the training data.
Instead of learning generalizable blendshapes, identity-specific animatable avatar methods~\cite{gao2022reconstructing, ma20243d, dhamo2024headgas, li2025rgbavatar} learn personalized blendshapes that capture appearance details and can be linearly combined with expression codes for efficient animation. Our method follows a similar idea and is most closely related to HeadGaS~\cite{dhamo2024headgas}, which learns Gaussian basis features for expression-dependent colors and opacities. However, HeadGaS optimizes identity-specific blending features from monocular videos, whereas our method generates dynamic blending bases for unseen identities from a single image. Moreover, HeadGaS keeps Gaussian positions fixed and models animation solely through colors and opacities. In contrast, our method uses the FLAME model~\cite{li2017learning} to deform Gaussian geometry and uses the generated blending bases to model expression-dependent appearance changes, resulting in more natural avatar dynamics.

% \vspace{-1.8pt}
\noindent\textbf{Generalizable Image-to-3D Reconstruction.}
Unlike traditional optimization-based methods, generalizable 3D reconstruction directly maps input images to 3D representations without test-time optimization. LRM~\cite{hong2024lrm} first demonstrates the ability of scaling up transformers and training data to learn 3D priors for feed-forward reconstruction. Subsequent works extend this paradigm to pose-free reconstruction~\cite{wang2024pf}, synthetic training data~\cite{xie2024lrm}, efficient 3DGS representations~\cite{zou2024triplane, tang2024lgm, zhang2024gs, xu2024grm}, and large-scale scene reconstruction~\cite{ziwen2025long,wang2026flashvggt}. However, these methods may struggle to recover regions occluded in the input images. To address this limitation, a line of work~\cite{xu2024dmv3d, cai2025baking,mi2026one4d, zhou2024diffgs, lin2025diffsplat} incorporate diffusion models for reconstruction. Our method follows this direction but focuses on animatable avatar reconstruction and dynamic appearance modeling rather than static reconstruction.

% \vspace{-20pt}
\vspace{-2pt}
\section{Method}
\label{sec:method}
\vspace{-3pt}

% \subsection{Overview}
Given a source image $\mathbf{I}_s$, we aim to reconstruct an animatable 3D Gaussian head avatar. Fig.~\ref{fig:overview} provides the framework of GauBlend. The canonical avatar diffusion model (CADM) takes $\mathbf{I}_s$ and 3DMM motions for noisy images as input and directly generates a UV-space canonical Gaussian avatar with a dynamic Gaussian blending basis for expression-dependent appearance modeling. During training, the canonical avatar is animated and rendered to denoise the images at each timestep, while one-step sampling is used at inference. During animation, given a driving expression code $\mathbf{e}$, dynamic Gaussian basis blending (DGBB) efficiently produces an animated Gaussian avatar $\mathcal{G}_a$ with expression-dependent appearance details, which is then rasterized under a given camera pose to produce the output image $\mathbf{I}_o$. The following subsections detail the canonical Gaussian avatar diffusion, the dynamic Gaussian basis blending, and the training strategy and objectives.
%utilize a transformer-based generative framework that  

\begin{figure*}[tb]
  \centering
  \includegraphics[width=0.89\linewidth]{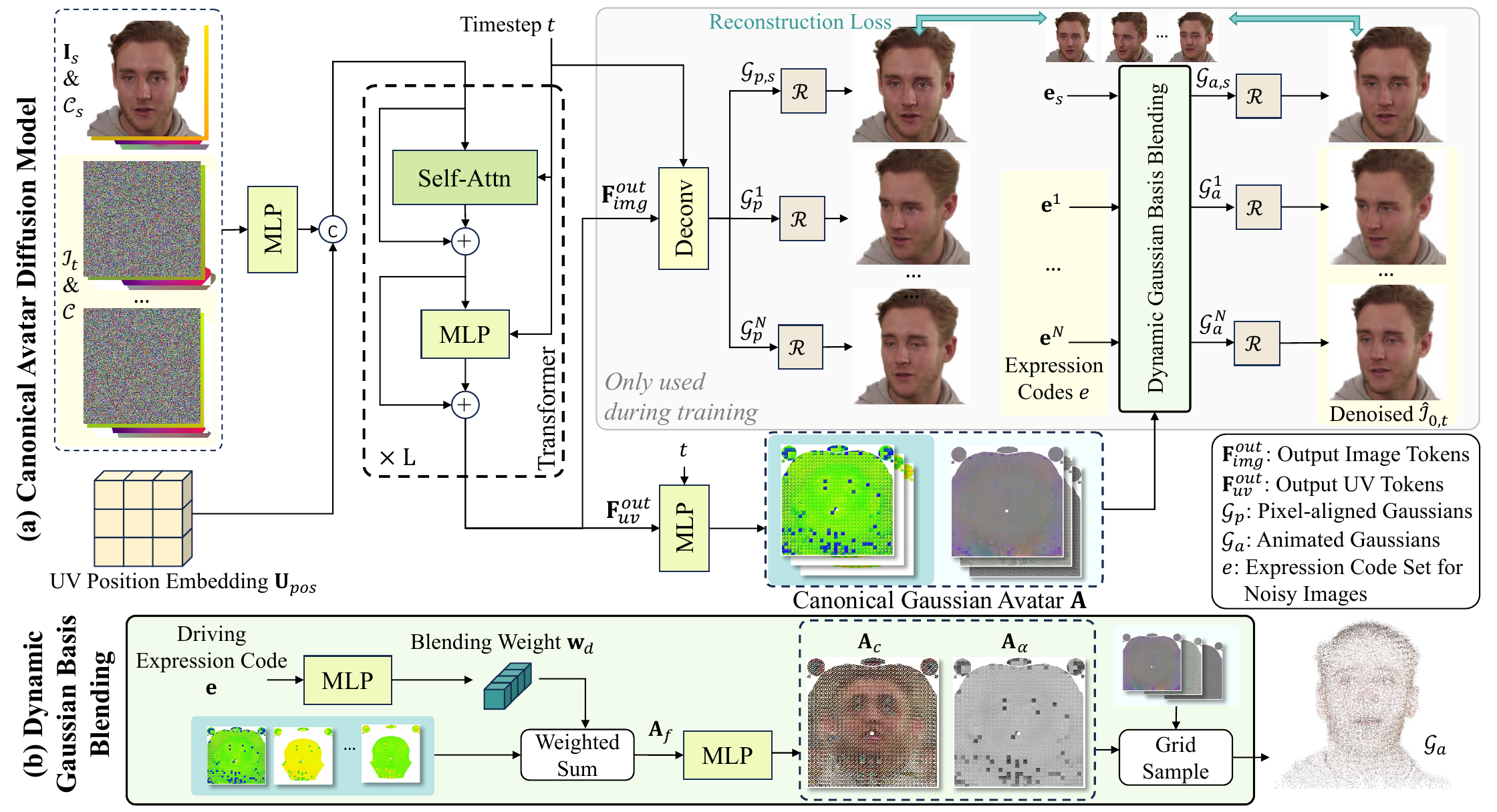}%{figures/diffusion_architecture.pdf}
  %{figures/overview_full_shadow.pdf}
  \vspace{-8pt}
  \caption{Illustration of (a) the canonical avatar diffusion model (CADM) and (b) the dynamic Gaussian basis blending module (DGBB).
  }
  \label{fig:diffusion}
  \vspace{-10pt}
\end{figure*}

\subsection{Canonical Gaussian Avatar Diffusion}
\label{sec:canonical_gaussian_diffusion}
%Previous approaches on one-shot feed-forward 3D animatable head avatars typically use a regression-based paradigm, directly predicting an avatar from the source image. Although they achieve promising results, their reconstruction quality is limited by the single input image and the regression-based architectures. To address these limitations, 
We propose a one-step canonical avatar diffusion model (CADM) that 
denoises images depicting novel poses and expressions with a single source image $\mathbf{I}_s$ to generate a UV-space canonical Gaussian avatar $\mathbf{A}$ in a single forward pass at inference. 
The architecture is shown in Fig.~\ref{fig:diffusion}a. Instead of predicting the noise at each timestep during training~\cite{ho2020denoising}, CADM directly predicts the canonical Gaussian avatar, following~\cite{cai2025baking, anciukevivcius2023renderdiffusion, xu2024dmv3d}. This enables direct supervision from 2D video frames by animating the predicted avatar under target poses and expressions, while preserving 3D consistency without requiring 3D ground-truth supervision. 
The canonical Gaussian avatar generation process is formulated as:
\begin{equation}
\mathbf{A}=\mathrm{CADM}(\mathbf{I}_s, \mathcal{C}_s;\mathcal{I}_t,\mathcal{C};t),
\end{equation}
where $\mathcal{I}_t=\{\mathbf{I}_t^i\}_{i=1}^N$ denotes a set of $N$ noisy images at timestep $t$, and $\mathcal{C}_s$ and $\mathcal{C}$ denote the conditioning signals for $\mathbf{I}_s$ and $\mathcal{I}_t$, respectively.

\noindent\textbf{Image, Camera, and Motion Conditioning.}
CADM aims to reconstruct the canonical Gaussian avatar from the source and noisy images at each diffusion timestep. As shown in Fig.~\ref{fig:overview}a, the source image $\mathbf{I}_s$ serves as the image condition, while auxiliary conditioning maps encode dense camera and motion information for the source and noisy images. For camera conditioning, we use Pl\"ucker coordinate~\cite{cai2025baking} to embed the camera views, producing camera maps $\mathbf{V}_s$ for $\mathbf{I}_s$ and $\mathcal{V}=\{\mathbf{V}^i\}_{i=1}^N$ for $\mathcal{I}_t$. For motion conditioning, following~\cite{taubner2025cap4d}, we use rasterized template meshes with motion-related vertex values. Specifically, an offline 3DMM estimator~\cite{qian2024gaussianavatars} estimates the FLAME~\cite{li2017learning} parameters for each image, which are then used to deform the FLAME mesh. For head pose conditioning, we assign the template vertex positions to the corresponding vertices. For fine-grained expression conditioning, we assign each vertex the displacement between its positions in the neutral- and tracked-expression meshes. This yields the pose maps $\mathbf{P}_s$ and $\mathcal{P}=\{\mathbf{P}^i\}_{i=1}^N$, and the expression maps $\mathbf{E}_s$ and $\mathcal{E}=\{\mathbf{E}^i\}_{i=1}^N$. The conditioning signals are therefore defined as $\mathcal{C}_s=\{\mathbf{V}_s, \mathbf{P}_s, \mathbf{E}_s\}$ and $\mathcal{C}=\{\mathcal{V}, \mathcal{P}, \mathcal{E}\}$. 
As shown in Fig.~\ref{fig:diffusion}a, we concatenate each image with its corresponding conditioning maps and encode the resulting inputs with an MLP for image tokenization.

\noindent\textbf{Canonical Gaussian Avatar Generation and Animation.}
%To allow efficient avatar animation at inference time, we model our canonical Gaussian avatar in the UV space of the FLAME mesh. 
To simplify the prediction of canonical Gaussian positions and their displacement during animation, we represent the canonical avatar in the FLAME mesh's UV space, which provides coarse head positions and enables the Gaussians to deform with the mesh during animation, following~\cite{kirschstein2024gghead}. Therefore, we use a learnable UV position embedding $\mathbf{U}_{pos}$ to query the input image tokens $\mathbf{F}_{img,t}^{in}$ through an $L$-layer transformer~\cite{peebles2023scalable}, and decode the output into the final UV-space canonical Gaussian avatar representation $\mathbf{A}$ with an MLP. The process is formulated as:
\begin{equation}
\mathbf{F}_{uv}^{out}, \mathbf{F}_{img}^{out} = \mathrm{Transformer}([\mathbf{U}_{pos}, \mathbf{F}_{img,t}^{in}],t), 
\end{equation}
\begin{equation}
\mathbf{A} = \mathrm{MLP}(\mathbf{F}_{uv}^{out}, t), 
\end{equation}
where $[\cdot,\cdot]$ denotes concatenation, $t$ is the diffusion timestep, and $\mathbf{F}_{uv}^{out}$ and $\mathbf{F}_{img}^{out}$ are the output UV and image tokens, respectively.
The canonical Gaussian avatar $\mathbf{A}$ comprises a scale map $\mathbf{A}_s$, a rotation map $\mathbf{A}_q$, a FLAME-relative position-offset map $\mathbf{A}_{\Delta\mu}$, and a dynamic blending-basis map $\mathbf{A}_b$. Given an expression code $\mathbf{e}$, the dynamic Gaussian basis blending module (DGBB) converts $\mathbf{A}$ into an animated Gaussian avatar $\mathcal{G}_{a}$:
\begin{equation}
\mathcal{G}_{a} = \mathrm{DGBB}(\mathbf{A}, \mathbf{e}). 
\label{eq:avatar_animation}
\end{equation}
Details of DGBB are provided in Sec.~\ref{subsec:blending}.

\noindent\textbf{Canonical Gaussian Avatar Supervision with Pixel-Aligned Gaussian Regularization.}
According to Eq.~\eqref{eq:avatar_animation}, given the expression codes $e=\{\mathbf{e}^i\}_{i=1}^N$ for the noisy images, we obtain animated Gaussian avatars $\{\mathcal{G}_{a}^i\}_{i=1}^N$. Rendering them from the corresponding camera poses yields denoised images $\hat{\mathcal{I}}_{0,t}$, which are supervised by the clean images $\mathcal{I}_{0}$ using
\begin{equation}
\mathcal{L}_{de} = \mathcal{L}_{rec}(\hat{\mathcal{I}}_{0,t}, \mathcal{I}_{0}), 
\end{equation}
where $\mathcal{L}_{rec}=\mathcal{L}_{2}+\mathcal{L}_{lpips}+\mathcal{L}_{id}$ is a reconstruction loss.
We also apply an $\mathcal{L}_2$ loss to the rendered alpha maps, denoted as $\mathcal{L}_{de}^{\alpha}$.
To improve reconstruction quality, we similarly animate and render $\mathbf{A}$ to reconstruct the source image $\mathbf{I}_{s}$ and $M$ randomly sampled images with novel views and motions, $\mathcal{I}_{nv}=\{\mathbf{I}_{nv}^j\}_{j=1}^M$. The resulting multi-view loss is:
\begin{equation}
\mathcal{L}_{mv} = \mathcal{L}_{rec}(\hat{\mathbf{I}}_{s,t}, \mathbf{I}_{s})+\mathcal{L}_{rec}(\hat{\mathcal{I}}_{nv,t}, \mathcal{I}_{nv}),
\label{eq:mv}
\end{equation}
where $\hat{\mathbf{I}}_{s,t}$ and $\hat{\mathcal{I}}_{nv,t}$ are the reconstructed source and novel-view images, respectively. We also similarly supervise their alpha maps, denoted as $\mathcal{L}_{mv}^{\alpha}$. %By combining the denoising loss and the multi-view loss, we can supervise the canonical Gaussian avatar generation.
%by obtaining the animated Gaussian avatars 
%obtain the animated Gaussian avatars for the source image and $M$ randomly sampled novel frames with novel motions and render them to the corresponding camera poses for reconstruction. 
%$\mathbf{e}_s$ and
%$\mathcal{G}_{a,s}$ and 

% However, we find that adding supervision solely to the renderings based on $\mathbf{A}$ cannot make them preserve appearance details from the source image, resulting in unfaithful reconstruction. 
% We attribute this to two reasons. First, the MLP-based image tokenizer is not well-trained to capture rich visual cues from the source image. Second, the noisy image tokens fail to denoise themselves with the appearance information from the source image tokens through all the transformer layers, making it difficult to provide the UV tokens with additional information from novel views and motions. 
However, supervision based solely on renderings from $\mathbf{A}$ may fail to preserve fine appearance details.
This is because the MLP-based image tokenizer may not extract sufficiently rich visual cues, and the noisy image tokens may not effectively exchange appearance information with the source image tokens through the transformer.
To address this issue, we regularize the output image tokens $\mathbf{F}_{img}^{out}$ by decoding them into pixel-aligned Gaussians~\cite{szymanowicz2024splatter} and supervise their renderings using $\mathbf{I}_{s}$ and $\mathcal{I}_{0}$. 
This encourages the image tokenizer to capture detailed appearance information and the noisy tokens to denoise effectively. 
We use self-attention layers to facilitate information exchange between the UV tokens and the image tokens in the transformer. Specifically, a deconvolution block~\cite{dumoulin2016guide} predicts pixel-aligned Gaussians for the source and noisy images, denoted by $\mathcal{G}_{p,s}$ and $\{\mathcal{G}_{p}^i\}_{i=1}^N$. Rendering them produces $\hat{\mathbf{I}}_{p,s}$ and $\hat{\mathcal{I}}_{p}=\{\hat{\mathbf{I}}_{p}^i\}_{i=1}^N$, which are supervised by a reconstruction loss:
%, thus improving the generation quality of both the canonical Gaussians and the pixel-aligned Gaussians. Specifically, we use a deconvolution block~\cite{dumoulin2016guide} to predict pixel-aligned Gaussians of the source and noisy images, which helps preserve the original image structure. The resulting Gaussians, $\mathcal{G}_{p,s}$ and $\{\mathcal{G}_{p}^i\}_{i=1}^N$, are rendered to the corresponding camera views, producing the reconstructed images $\hat{\mathbf{I}}_{p,s}$ and $\hat{\mathcal{I}}_{p}=\{\hat{\mathbf{I}}_{p}^i\}_{i=1}^N$. We then supervise the pixel-aligned Gaussian generation as:
\begin{equation}
\mathcal{L}_{reg} = \mathcal{L}_{rec}'(\hat{\mathbf{I}}_{p,s}, \mathbf{I}_{s})+\mathcal{L}_{rec}'(\hat{\mathcal{I}}_{p}, \mathcal{I}_{0}), 
\end{equation}
where $\mathcal{L}_{rec}'=\mathcal{L}_{2}+\mathcal{L}_{lpips}$. We also apply an $\mathcal{L}_{2}$ loss on the rendered alpha maps, denoted as $\mathcal{L}_{reg}^{\alpha}$.
Since the pixel-aligned Gaussians are used only for regularization, we directly predict their colors and opacities instead of blending bases.
% Note that instead of predicting Gaussian blending basis, we directly predict the color and opacity attribute values for pixel-aligned Gaussians, as we do not animate them. We predict them only for regularization to improve the reconstruction fidelity of the canonical Gaussians. 
% Finally, we regularize the position maps between the animated Gaussians $\mathcal{G}_{a}$ and the corresponding pixel-aligned Gaussians $\mathcal{G}_{p}$ to be close to each other at the corresponding camera poses for the source and noisy images, as they are the 3D reconstruction of the same image and should be identical ideally. Specifically, we substitute the color value with the position value for each group of Gaussians, and render them with the corresponding camera views. Thus, we obtain the position maps of the pixel-aligned Gaussians $\hat{\mathbf{I}}_{p,s}^{\mu}$ and $\hat{\mathcal{I}}_{p}^{\mu}$, and the position maps of the animated Gaussians $\hat{\mathbf{I}}_{s,t}^{\mu}$ and $\hat{\mathcal{I}}_{0,t}^{\mu}$. We regularize them as:
Finally, we regularize the positions of the animated Gaussian avatar and the corresponding pixel-aligned Gaussians, which represent the same images and should therefore agree in 3D. Replacing Gaussian colors with position values and rendering from the corresponding views yields the position maps $\hat{\mathbf{I}}_{p,s}^{\mu}$, $\hat{\mathcal{I}}_{p}^{\mu}$, $\hat{\mathbf{I}}_{s,t}^{\mu}$ and $\hat{\mathcal{I}}_{0,t}^{\mu}$. We impose
\begin{equation}
\begin{split}
\mathcal{L}_{reg}^{\mu} &= \mathcal{L}_{2}(\hat{\mathbf{I}}_{p,s}^{\mu}, sg[\hat{\mathbf{I}}_{s,t}^{\mu}])+\mathcal{L}_{2}(\hat{\mathcal{I}}_{p}^{\mu}, sg[\hat{\mathcal{I}}_{0,t}^{\mu}])\\
&+ \beta (\mathcal{L}_{2}(sg[\hat{\mathbf{I}}_{p,s}^{\mu}], \hat{\mathbf{I}}_{s,t}^{\mu})+\mathcal{L}_{2}(sg[\hat{\mathcal{I}}_{p}^{\mu}], \hat{\mathcal{I}}_{0,t}^{\mu})),
\end{split}
\end{equation}
where $sg[\cdot]$ denotes stop-gradient, and $\beta$ is a loss weight.

%cannot make it preserve appearance details from the source image 
%However, the generated avatar cannot fully preserve appearance details from the source image. We assume that the input image tokens "cannot well encode the source image feature, the noisy image tokens cannot be well denoised itself, thus cannot provide novel view and pose information for the UV position embedding to combine" 
%, which contain features from noisy images, cannot 

%Although the input image tokens contain dense conditioning signals to denoise the noisy images, 

% Previous image-to-3D Gaussian reconstruction methods~\cite{cai2025baking, tang2024lgm, xu2024grm, zhang2024gs, szymanowicz2024splatter} typically predict pixel-aligned Gaussians from input image tokens. Although they can effectively reconstruct static 3D objects and scenes, it is non-trivial to  

% \noindent\textbf{Gaussian Avatar Animation.}

\noindent\textbf{One-step Inference.}
Because CADM directly predicts the canonical Gaussian avatar $\mathbf{A}$ at each timestep during training, it supports feed-forward one-step inference. Given the source image, its conditions, pure noise, and randomly sampled camera and motion conditions, CADM directly generates $\mathbf{A}$. Experiments show that one-step generation outperforms multi-step sampling (See Supplementary Material); we therefore adopt it for both high fidelity and efficiency.
%We find one-step inference performs the best (shown in Experiments), and is the most efficient.

% \begin{figure}[tb]
%   \centering
%   \includegraphics[width=0.99\linewidth]{figures/basis_blending.pdf}
%   %{figures/overview_full_shadow.pdf}
%   % \vspace{-8pt}
%   \caption{Illustration of dynamic Gaussian basis blending. The dynamic Gaussian blending basis are linearly blended with a compressed driving expression code and decoded into Gaussian colors and opacities. Together with other canonical Gaussian attributes, they construct an animated Gaussian avatar.
%   }
%   \label{fig:basis_blending}
%   % \vspace{-10pt}
% \end{figure}

%%%%%%%%%%%%%%%%%%%%%%%% HDTF
\begin{table*}[tb]
  % \vspace{-10pt}
  %\begin{center}
  \caption{Quantitative comparison with state-of-the-art methods on the HDTF dataset~\cite{zhang2021flow}.}
  \label{table:sota_hdtf}
  \vspace{-10pt}
  \centering
  \resizebox{0.85\textwidth}{!}{ %
  \begin{tabular}{l|ccccccc|ccc} %l}
  \toprule[1.2pt]
    \multirow{2}{*}{Method}  &  \multicolumn{7}{c|}{Self-reenactment}  &  \multicolumn{3}{c}{Cross-identity Reenactment} \\
    %\cline{2-7} \cline{8-13}
    
             & PSNR $\uparrow$  & SSIM $\uparrow$   & LPIPS $\downarrow$ & CSIM $\uparrow$  & AKD $\downarrow$ & AED $\downarrow$ & %APD $\downarrow$
             APD $\downarrow$
             & CSIM $\uparrow$  & AED $\downarrow$ & APD $\downarrow$\\
    \midrule
    Portrait4D-v2~\cite{deng2024portrait4d_eccv}
    & 18.92 & 0.8116 & 0.2146 & 0.9244 %0.8622 
    & 2.7266 & 0.3437 & 1.1959
    &  0.8652 %0.8636 
    & 0.7358 &  1.3739
    \\
    
    GAGAvatar~\cite{chu2024generalizable}
    & {19.97} & {0.8280} & {0.2025} & 0.9442 %\textbf{0.8899} 
    & 2.6813 & {0.3056} & {0.5840}
    & 0.8626 %\textbf{0.8996} 
    & 0.7280 & \textbf{1.2229}
    \\

    LAM~\cite{he2025lam}
    & 19.33 & 0.8012 & 0.2387 & 0.8856 %0.7529 
    & 3.3250 & 0.4177 & 0.9274
    & 0.8274 %0.7650 
    &  0.7746 & 1.6436
    \\

    UIKA~\cite{wu2026uika}
    &  19.04 & 0.8075 & 0.2196 & 0.9206 %0.8480
    & 3.0720 & 0.3453 & 0.7378
    & 0.8473 % 0.8577 
    & 0.7331 & 1.4195
    \\

    FlexAvatar~\cite{kirschstein2026flexavatar}
    & 19.69 & 0.8100 & 0.2197 & 0.9328 %0.8003 
    & {2.5101} & 0.3069 & 0.6086
    & 0.8410  %0.7934 
    & \textbf{0.6508} & 1.3043
    \\
    
    Ours 
    & \textbf{23.70} & \textbf{0.8608} & \textbf{0.1816} & \textbf{0.9557} %{0.8628} 
    & \textbf{2.2619} & \textbf{0.3054} & \textbf{0.5285}  
    % & \textbf{0.8759} %{0.8670}  
    % & {0.7262} & {1.3705}
    & \textbf{0.8737}
    & 0.7174 & 1.3112
    \\ 

    \bottomrule[1.2pt]
  \end{tabular}
  % \vspace{-8pt}
  }
  \vspace{-8pt}
  
\end{table*}

\subsection{Dynamic Gaussian Basis Blending} % for Efficient Dynamic Appearance Modeling}
\label{subsec:blending}
We model expression-dependent appearance for natural, fast avatar animation by linearly blending the dynamic Gaussian blending basis map $\mathbf{A}_b$ with a driving expression code $\mathbf{e}$ and decoding the result into Gaussian colors and opacities. Combined with the remaining attributes from $\mathbf{A}$, they form the animated avatar $\mathcal{G}_a$. Fig.~\ref{fig:diffusion}b illustrates the dynamic Gaussian basis blending module (DGBB).

\noindent\textbf{Expression-based Basis Blending.}
Defined in a UV space of size $K$, the dynamic blending-basis map $\mathbf{A}_b \in \mathbb{R}^{K\times K\times d_b(T+1)}$ comprises a base $\mathbf{A}_b^0$ and $T$ expression-related offsets $\{\Delta \mathbf{A}_b^k\}_{k=1}^T$, each of dimension $d_b$. The base $\mathbf{A}_b^0$ represents the base head appearance, while the offsets model expression-dependent changes and are defined only in the facial UV region to avoid unnecessary modifications elsewhere. Given a driving expression code $\mathbf{e}\in \mathbb{R}^{d_e}$, an MLP maps it to expression-dependent blending weights $\mathbf{w}_d=[w_d^1, w_d^2, ..., w_d^T] \in \mathbb{R}^{T}$, which control the combination of the blending basis. We perform linear blending as:
\begin{equation}
\mathbf{A}_f=\mathbf{A}_b^0 + \sum_{k=1}^T w_d^k\Delta \mathbf{A}_b^k,
\end{equation}
where $\mathbf{A}_f\in \mathbb{R}^{K\times K\times d_b}$ denotes the blended feature map. By combining the basis features with expression-dependent weights, $\mathbf{A}_f$ captures dynamic appearance details, such as wrinkles, for expression $\mathbf{e}$. An MLP then decodes $\mathbf{A}_f$ into Gaussian color and opacity maps $\mathbf{A}_{c}$ and $\mathbf{A}_{\alpha}$:
\begin{equation}
\mathbf{A}_{c}, \mathbf{A}_{\alpha}=\mathrm{MLP}(\mathbf{A}_f+\mathbf{U}_{pos}^{sin}),
\end{equation}
where $\mathbf{A}_{c} \in \mathbb{R}^{K\times K\times 3}$, $\mathbf{A}_{\alpha}\in \mathbb{R}^{K\times K\times 1}$, and $\mathbf{U}_{pos}^{sin}$ denotes the sinusoidal positional encoding of the UV coordinates.

\noindent\textbf{Animated Gaussian Avatar Construction.}
We combine the UV-space Gaussian color and opacity maps with the remaining attribute maps in $\mathbf{A}$ to form a complete Gaussian representation. Because $\mathbf{A}_{\Delta\mu}$ specifies offsets from the FLAME mesh surface, we obtain the 3D Gaussian position map $\mathbf{A}_{\mu}$ by adding these position offsets to the mesh deformed by the driving motion. This yields the animated Gaussian maps $\mathbf{A}_a$: 
\begin{equation}
\mathbf{A}_{a}=[\mathbf{A}_{c}, \mathbf{A}_{\alpha}, \mathbf{A}_{s}, \mathbf{A}_{q}, \mathbf{A}_{\mu}],
\end{equation}
where $[\cdot,\cdot]$ denotes concatenation. We then construct the animated Gaussian avatar $\mathcal{G}_a$ by grid sampling:
\begin{equation}
\mathcal{G}_a=\mathrm{Grid\_Sample}(\mathbf{A}_a, \mathcal{S}),
\end{equation}
where $\mathcal{S}$ denotes the valid UV sampling positions.

\subsection{Training Strategy and Objectives}

\noindent\textbf{Two-stage Training.}
To achieve one-shot generalizable avatar reconstruction, we train on large-scale monocular video datasets. However, their limited viewpoint and head-pose diversity can cause severe distortions for views far from those seen during training. Following~\cite{zhao2026one}, we therefore use a pretrained 3D GAN inversion model~\cite{an2023panohead, bilecen2024dual} to provide pseudo-multi-view supervision. Specifically, we generate randomly sampled novel views of the source image $\mathbf{I}_s$ and include them in the novel-view set $\mathcal{I}_{nv}$ from Eq.~\eqref{eq:mv}. However, the inversion model may produce unnatural head geometry at certain viewpoints, limiting reconstruction quality. To mitigate these artifacts, we adopt a two-stage training strategy inspired by LCA~\cite{li2026large}. In the first stage, we train on large-scale monocular videos with pseudo-multi-view supervision, enabling generalization to unseen identities and coarse 3D geometry reconstruction. In the second stage, we remove the pseudo-multi-view supervision and incorporate real multi-view videos for more accurate 3D supervision. To avoid overfitting to the relatively small multi-view datasets, we jointly sample data from both monocular and multi-view datasets. This strategy combines the generalizability of monocular-video training with the high 3D reconstruction quality provided by real multi-view supervision.

\noindent\textbf{Training Objectives.}
In addition to the losses described in Sec.~\ref{sec:canonical_gaussian_diffusion}, we constrain the Gaussian position offsets from the mesh as:
\begin{equation}
\mathcal{L}_{\Delta\mu}=\max(||\Delta \mathbf{A}_{\Delta\mu}||_2-\epsilon, 0),
\end{equation}
where $\mathbf{A}_{\Delta\mu}$ is the canonical Gaussian position-offset map and $\epsilon$ is a threshold. The total training objective is:
\begin{equation}
\begin{split}
\mathcal{L}&=\mathcal{L}_{de}+\mathcal{L}_{de}^{\alpha}+\mathcal{L}_{mv}+\mathcal{L}_{mv}^{\alpha}\\
&+\lambda \mathcal{L}_{reg}+\lambda^{\alpha} \mathcal{L}_{reg}^{\alpha}+\mathcal{L}_{reg}^{\mu}
+\mathcal{L}_{\Delta\mu},
\end{split}
\end{equation}
where $\lambda$ and $\lambda^{\alpha}$ are loss weights. 
% only replaced figure "results/exp_samecrossid_qualitative.pdf" with "results/exp_samecrossid_qualitative_refinemisaligned.pdf" to remove the gray artifacts at row 4 between column 2 and 3.

\section{Experiments}
\label{sec:exp}

\subsection{Experimental Setup}
\noindent\textbf{Datasets.}
We use VFHQ~\cite{xie2022vfhq} for first-stage training, and further use NeRSemble-v2~\cite{kirschstein2023nersemble} and Ava-256~\cite{martinez2024codec} for second-stage training. For VFHQ, we crop and remove the background of the frames and estimate the camera poses following~\cite{an2023panohead, bilecen2024dual}. For NeRSemble-v2 and Ava-256, we crop the frames following~\cite{chu2024generalizable}. All frames are resized to $512\times 512$, and FLAME parameters are estimated using the head tracker from~\cite{qian2024gaussianavatars}. For evaluation, we use the VFHQ test split and 20 videos from HDTF~\cite{zhang2021flow}, following prior works~\cite{chu2024generalizable, he2025lam, zhao2026one}. 
%We further evaluate the 3D reconstruction performance on 30 multi-view videos from 10 identities from NeRSemble-v2, which we hold out during training.

\noindent\textbf{Evaluation Metrics.}
We evaluate the performance of self- and cross-identity reenactment, following prior works~\cite{chu2024generalizable, he2025lam}. For self-reenactment, we measure reconstruction quality using PSNR, SSIM, and LPIPS~\cite{zhang2018unreasonable}, and identity similarity using the cosine similarity (CSIM) of identity embeddings from~\cite{amos2016openface} between the source and rendered images. We evaluate motion transfer using average keypoint distance (AKD)~\cite{bulat2017far}, average expression distance (AED)~\cite{retsinas20243d} and average pose distance (APD)~\cite{lugaresi2019mediapipe}, which measure differences in facial keypoints, expression coefficients, and pose coefficients between the driving and rendered images. For cross-identity reenactment, we report CSIM, AED, and APD because ground-truth frames are unavailable.

\begin{figure*}[tb]
    \centering
  \includegraphics[width=0.9\linewidth]{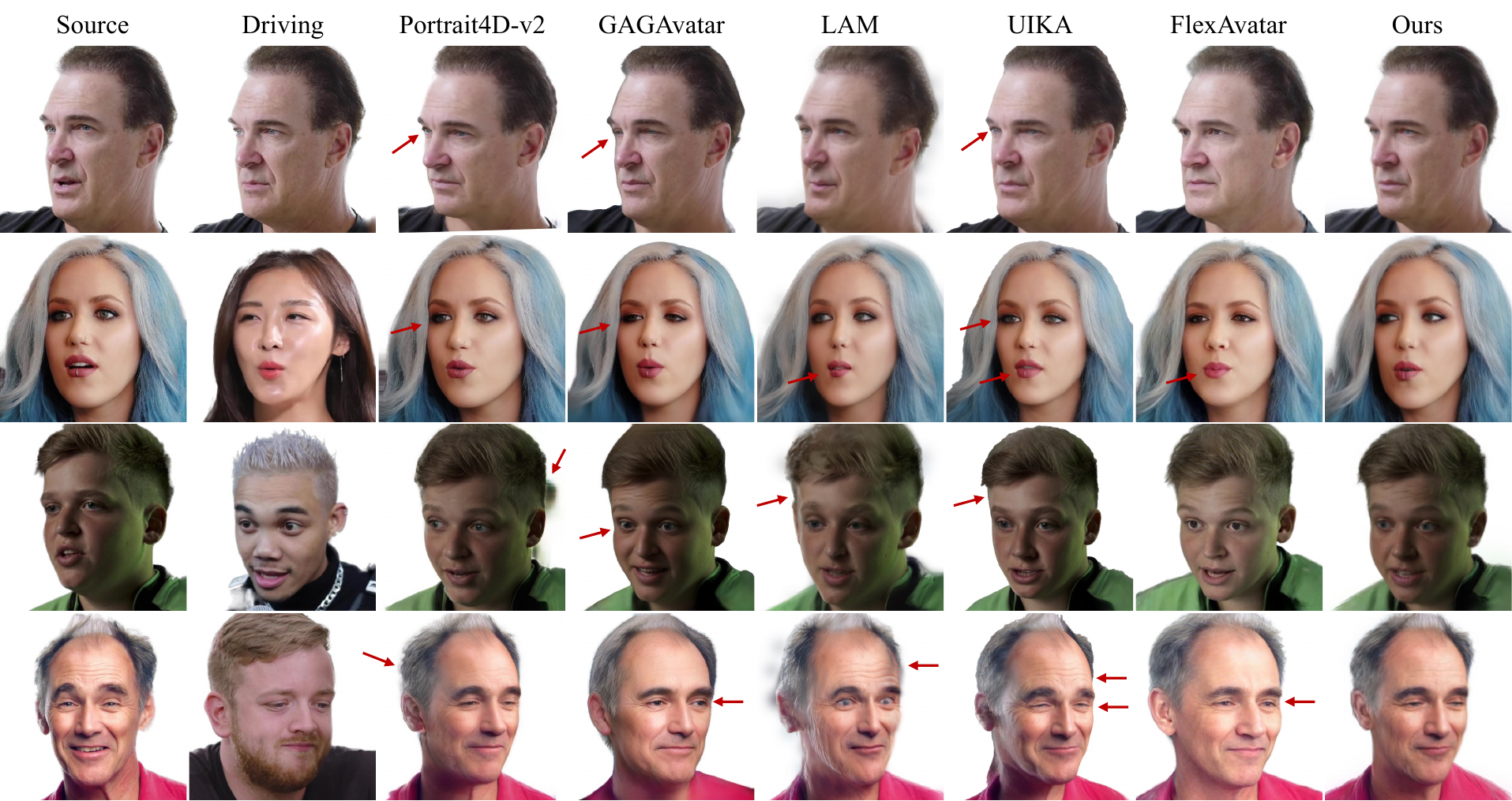}
  %{results/exp_crossid_qualitative.pdf}

  \vspace{-9pt}
  \caption{Qualitative comparison with state-of-the-art methods for self- and cross-identity reenactment on the VFHQ dataset~\cite{xie2022vfhq}. Our method follows the driving motion accurately while preserving source identity.
  % (\ie, Portrait4D-v2~\cite{deng2024portrait4d_eccv}, GAGAvatar~\cite{chu2024generalizable}, LAM~\cite{he2025lam}, UIKA~\cite{wu2026uika} and FlexAvatar~\cite{kirschstein2026flexavatar})
  }
  \vspace{-4pt}
  \label{fig:sota_cmp_crossid}
\end{figure*}

%%%%%%%%%%%%%%%%%%%%%%%% VFHQ
\begin{table*}[!tb]
\caption{Quantitative comparison with state-of-the-art methods on the VFHQ dataset~\cite{xie2022vfhq}.}
  \label{table:sota_vfhq}
  % \vspace{-14pt}
  \vspace{-10pt}
  %\begin{center}
  \centering
  \resizebox{0.85\textwidth}{!}{ %
  \begin{tabular}{l|ccccccc|ccc} %l}
  \toprule[1.2pt]
    \multirow{2}{*}{Method}  &  \multicolumn{7}{c|}{Self-reenactment}  &  \multicolumn{3}{c}{Cross-identity Reenactment} \\
    %\cline{2-7} \cline{8-13}
    
             & PSNR $\uparrow$  & SSIM $\uparrow$   & LPIPS $\downarrow$ & CSIM $\uparrow$  & AKD $\downarrow$ & AED $\downarrow$ & %APD $\downarrow$
             APD $\downarrow$
             & CSIM $\uparrow$  & AED $\downarrow$ & APD $\downarrow$\\
    \midrule
    Portrait4D-v2~\cite{deng2024portrait4d_eccv}
    & 15.56 & 0.7316 & 0.2785 
    %& 0.8010 
    & 0.9099
    & 3.2247 &  0.3537 & 1.5891
    % & 0.7435 %0.6714 
    % & 0.6579 &  2.1437
    & 0.7293 & 0.6563 & 2.2309 % v3
    \\
    
    GAGAvatar~\cite{chu2024generalizable}
    & 17.87 & {0.7766} & {0.2358} 
    %& \textbf{0.8409} 
    & 0.9276 
    & {2.9079} & \textbf{0.3152} & {0.8561}
    % & 0.7566 %0.6653 
    % & 0.6308 & 1.7146
    & 0.7583 & 0.6507 & 1.7428 % v3
    \\

    LAM~\cite{he2025lam}
    & {18.23} & 0.7534 & 0.2685 
    %& 0.6834 
    & 0.9031
    & 3.4684 & 0.4376 & 1.3065
    % &  0.7255 %0.5539 
    % & 0.7212 & 3.2246
    & 0.7048 & 0.7297 & 3.0536 % v3
    \\

    UIKA~\cite{wu2026uika}
    & 17.47 & 0.7563 & 0.2496 
    % & 0.7746
    & 0.9164
    & 3.7413 & 0.3704 & 1.2448
    % & \textbf{0.7603} %0.6722 
    % & 0.6755 & 1.9979
    & \textbf{0.7608} & 0.6778 & 2.2214 % v3
    \\

    FlexAvatar~\cite{kirschstein2026flexavatar}
    & 18.03 & 0.7571 & 0.2532 
    % & 0.7375 
    & 0.9171
    & 3.0393 & 0.3303 & 0.9084
    % & 0.7375 %0.6150 
    % & \textbf{0.5872} & \textbf{1.6236}
    & 0.7119 & \textbf{0.5825} & \textbf{1.6046} % v3
    \\
    
    Ours 
    & \textbf{20.41} & \textbf{0.7988} & \textbf{0.2213} 
    % & {0.8026} 
    & \textbf{0.9373}
    & \textbf{2.5571} & {0.3213} & \textbf{0.8207}   
    % & {0.7567} %\textbf{0.6739} 
    % & 0.6571 & 2.6395
    % & 0.7475 &  0.6455 & 2.1278 % driving motion
    & 0.7401 & 0.6436 & 2.1210 % v3: driving motion 
    \\ 

    \bottomrule[1.2pt]
  \end{tabular}
  }
  \vspace{-8pt}
  
\end{table*}

\begin{table}[tb]
\caption{Animation speed measured in FPS compared with methods that model dynamic appearances. The time for source avatar reconstruction and driving motion extraction that can be computed beforehand is excluded.
  }
  \vspace{-10pt}
  \label{table:speed_fps}
  \centering
  \resizebox{1
  \linewidth}{!}{
  \begin{tabular}{l|cccc} 
  \toprule[1.2pt]
  Method  &  
    
    Portrait4D-v2~\cite{deng2024portrait4d_eccv}
      & 
    
    GAGAvatar~\cite{chu2024generalizable}
      &

    FlexAvatar~\cite{kirschstein2026flexavatar}
    &

    Ours 
  
  \\
             
    \midrule
    FPS & 18.86 & 59.50 & 50.39 & \textbf{114.47} 
    \\ 

    \bottomrule[1.2pt]
  \end{tabular}
  }
  \vspace{-8pt}
  
\end{table}

\subsection{Comparison with State-of-the-art Methods}
We compare GauBlend with recent open-source state-of-the-art methods for one-shot feed-forward animatable head avatar reconstruction, including Portrait4D-v2~\cite{deng2024portrait4d_eccv}, GAGAvatar~\cite{chu2024generalizable}, LAM~\cite{he2025lam}, UIKA~\cite{wu2026uika}, and FlexAvatar~\cite{kirschstein2026flexavatar}. In particular, LAM and UIKA do not model dynamic appearance changes, while Portrait4D-v2, GAGAvatar and FlexAvatar model them by injecting expression codes into a 3D avatar decoder or applying a 2D neural renderer to coarse renderings.

\noindent\textbf{Quantitative Comparison.}
Tab.~\ref{table:sota_hdtf} and Tab.~\ref{table:sota_vfhq} present the quantitative comparison results with state-of-the-art methods on HDTF~\cite{zhang2021flow} and VFHQ~\cite{xie2022vfhq}. Our method achieves the best overall performance in self-reenactment and delivers strong results in cross-identity reenactment. It significantly outperforms existing approaches in reconstruction and identity preservation, as measured by PSNR, SSIM, LPIPS, and CSIM, and generally achieves the best driving accuracy in self-reenactment according to AKD, AED, and APD. In particular, our method substantially outperforms LAM~\cite{he2025lam} and UIKA~\cite{wu2026uika}, which do not model dynamic appearance changes. This highlights the importance of such modeling for natural avatar motion. Our method also achieves competitive driving accuracy compared with approaches that model dynamic appearance (\ie, Portrait4D-v2~\cite{deng2024portrait4d_eccv}, GAGAvatar~\cite{chu2024generalizable}, FlexAvatar~\cite{kirschstein2026flexavatar}). 
This demonstrates the effectiveness of dynamic Gaussian blending bases generated by CADM.
For cross-identity reenactment, our method achieves the highest CSIM on HDTF, confirming robust identity preservation across datasets.  
%our method achieves slightly inferior driving accuracy. 
% For driving accuracy, our method produces slightly inferior AED and APD metrics, which might be due to the limited tracking accuracy of our head tracker. However, our AED metric consistently outperforms methods without dynamic appearance modeling and remains strong compared to methods modeling dynamic appearance with heavier networks.
Although our AED and APD are slightly lower, likely due to limitations in head-tracking accuracy, our AED consistently surpasses that of methods without dynamic appearance modeling and remains competitive with methods using heavier dynamic-appearance networks.

%although our pose accuracy is limited by the FLAME tracking accuracy, our method still achieves the best AED on HDTF, confirming that the generated Gaussian blending bases are robust under diverse expressions.
%, which leads to a slightly inferior APD metric, it still achieves the 

\begin{figure*}[tb]
    \centering
   \includegraphics[width=0.99\linewidth]{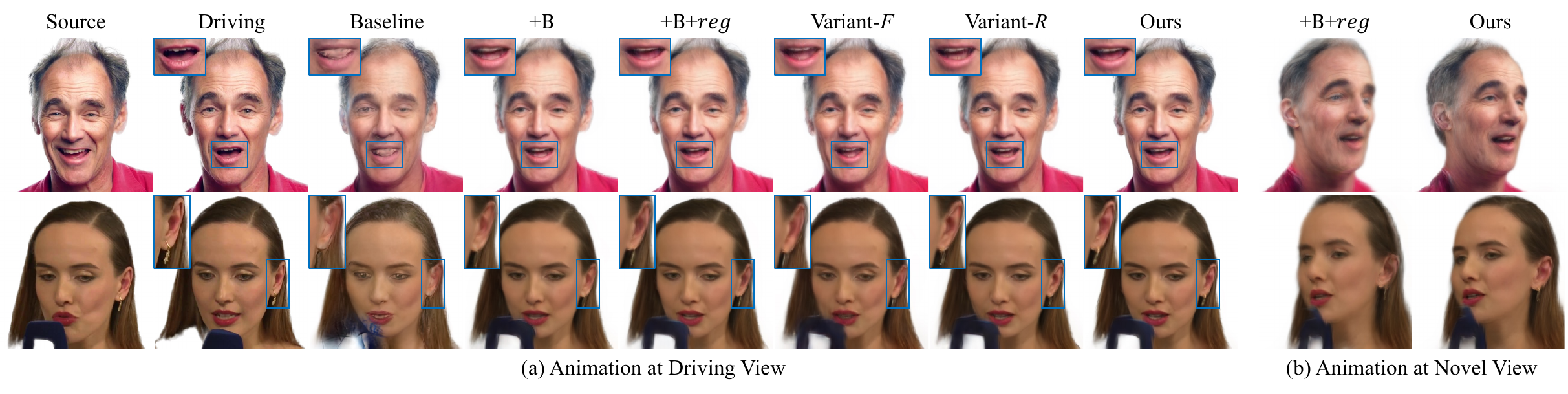}
   \vspace{-10pt}
  \caption{Qualitative ablation study on the VFHQ dataset~\cite{xie2022vfhq}. Our full method best reconstructs the source appearances, producing expression-dependent appearance details while achieving high 3D visual consistency.
  }

  \label{fig:ablation_design}
  \vspace{-8pt}
\end{figure*}

% \begin{figure}[tb]
%     \centering
%    \includegraphics[width=1\linewidth]{results/exp_abla_basis_vis.pdf}
%    % \vspace{-15pt}
%   \caption{Visualization of the learned Gaussian blending bases. Only by altering the weights to blend the Gaussian blending bases to change the Gaussian colors and opacities while keeping other Gaussian attributes fixed, the rendered avatars resemble diverse expressions.
%   }

%   \label{fig:ablation_bases}
%   % \vspace{-15pt}
% \end{figure}

\noindent\textbf{Qualitative Comparison.}
Fig.~\ref{fig:sota_cmp_crossid} presents qualitative comparisons for self- and cross-identity reenactment on VFHQ~\cite{xie2022vfhq}. Our method best follows the driving motion while preserving the source identity. It also produces expression-dependent appearance details absent from the source image, such as adding forehead wrinkles (row 3) and removing them (row 4). In contrast, LAM~\cite{he2025lam} and UIKA~\cite{wu2026uika} directly use the source appearance for all expressions, resulting in stiff animations (rows 3 and 4). Although other methods model dynamic appearance, they often fail to reproduce fine motions around the eyes and mouth. While FlexAvatar~\cite{kirschstein2026flexavatar} achieves strong quantitative motion transfer results for cross-identity reenactment, it fails to preserve the source identity in a feed-forward manner, such as the skin tone in row 3, thereby reducing animation quality. More qualitative results are shown in the Supplementary Material.

\noindent\textbf{Animation Speed.}
Tab.~\ref{table:speed_fps} compares our animation speed with dynamic-appearance methods~\cite{deng2024portrait4d_eccv, chu2024generalizable, kirschstein2026flexavatar} on an NVIDIA RTX 3090. Their heavy networks for generating expression-dependent appearance limit animation speed. In contrast, our explicit dynamic Gaussian blending basis enables efficient blending, achieving 114 FPS and substantially outperforming prior methods, highlighting its potential for real-time applications.
%All results exclude the time for source avatar reconstruction and driving motion extraction that can be calculated beforehand. We take an average over 100 frames.

% %%%%%%%%%%%%%%%%%%%%%%%% NeRSemble
% \begin{table*}[!tb]
%   % \vspace{-10pt}
%   %\begin{center}
%   \centering
%   \resizebox{0.85\textwidth}{!}{ %
%   \begin{tabular}{l|ccccccc|ccc} %l}
%   \toprule[1.2pt]
%     \multirow{2}{*}{Method}  &  \multicolumn{7}{c|}{Self-reenactment}  &  \multicolumn{3}{c}{Cross-identity Reenactment} \\
%     %\cline{2-7} \cline{8-13}
    
%              & PSNR $\uparrow$  & SSIM $\uparrow$   & LPIPS $\downarrow$ & CSIM $\uparrow$  & AKD $\downarrow$ & AED $\downarrow$ & %APD $\downarrow$
%              APD $\downarrow$
%              & CSIM $\uparrow$  & AED $\downarrow$ & APD $\downarrow$\\
%     \midrule
%     Portrait4D-v2~\cite{deng2024portrait4d_eccv}
%     & 
%     & 
%     \\
    
%     GAGAvatar~\cite{chu2024generalizable}
%     & 
%     & 
%     \\

%     LAM~\cite{he2025lam}
%     & 
%     & 
%     \\

%     UIKA~\cite{wu2026uika}
%     & 
%     & 
%     \\

%     FlexAvatar~\cite{kirschstein2026flexavatar}
%     & 
%     & 
%     \\
    
%     Ours 
%     &    
%     & 
%     \\ 

%     \bottomrule[1.2pt]
%   \end{tabular}
%   }
%   % \vspace{-7pt}
%   \caption{Quantitative comparison with state-of-the-art methods on the NeRSemble dataset~\cite{kirschstein2023nersemble}.}
%   \label{table:sota_NeRSemble}
%   % \vspace{-14pt}
% \end{table*}

\subsection{Ablation Study}
\label{subsec:ablation}

We conduct ablation studies to analyze our method, with results presented in Tab.~\ref{table:ablation_design}, Fig.~\ref{fig:ablation_design} and Fig.~\ref{fig:teaser}. %Tab.~\ref{table:ablation_sampling_steps}, Fig.~\ref{fig:ablation_design} and Fig.~\ref{fig:teaser}.

\noindent\textbf{Dynamic Gaussian Blending Bases.}
As shown in Tab.~\ref{table:ablation_design} and Fig.~\ref{fig:ablation_design}a, directly predicting fixed Gaussian colors and opacities with the diffusion model (Baseline) yields poor reconstruction quality, likely because static appearance representation cannot adapt to facial dynamics. Incorporating dynamic Gaussian blending bases into the canonical avatar and blending them during animation (+$\mathrm{B}$) enables expression-dependent appearance modeling, substantially improving reconstruction quality on dynamic videos. We further visualize the learned bases in Fig.~\ref{fig:teaser}b %{fig:ablation_bases} 
by applying PCA and traversing along the principal component directions~\cite{harkonen2020ganspace}. Varying only the blending weights\textemdash while keeping all other Gaussian attributes, including positions, fixed\textemdash produces avatars with diverse expressions. This demonstrates that the learned bases capture rich dynamic appearance details, which are important for natural expression modeling.
%Only by altering the weights to blend the Gaussian blending bases to change the Gaussian colors and opacities while keeping other Gaussian attributes (including the position) fixed, the rendered avatars resemble diverse expressions. This confirms that our generated Gaussian blending bases contain diverse dynamic appearance details and highlights the importance of dynamic appearances for natural expression modeling.

\noindent\textbf{Pixel-Aligned Gaussian Regularization.}
Tab.~\ref{table:ablation_design} shows that pixel-aligned Gaussian regularization (+$reg$) further improves animated avatar quality. As illustrated in Fig.~\ref{fig:ablation_design}a, the improvement is particularly evident in local details, such as the mouth interior and the ear, which are difficult to reconstruct faithfully without fine-grained regularization.

\noindent\textbf{Diffusion Training.}
We evaluate diffusion training (+$\mathrm{B}$+$reg$) against two variants: fixed maximum-noise training and regression. Fixed maximum-noise training (Variant-\textit{F}) adds only the highest noise level during training. Although this matches one-step inference, it performs worse (see Tab.~\ref{table:ablation_design} and Fig.~\ref{fig:ablation_design}a), likely because learning from nearly pure-noise images is difficult. In contrast, diffusion training progressively teaches the model to denoise images across noise levels, which eases the training process. Regression-based training (Variant-\textit{R}) does not denoise novel images and relies solely on the source image, preventing it from exploiting additional information from denoised views and resulting in inferior performance.

\noindent\textbf{Two-Stage Training.}
We evaluate two-stage training (Ours) against the model trained with the same number of iterations but without using the real multi-view data in the second stage (+$\mathrm{B}$+$reg$). As shown in Fig.~\ref{fig:ablation_design}b, real multi-view videos correct the unnatural head shapes introduced by the pseudo-multi-view supervision from 3D GAN inversion~\cite{an2023panohead, bilecen2024dual}, thereby improving 3D consistency. Tab.~\ref{table:ablation_design} and Fig.~\ref{fig:ablation_design}a further show that fine-tuning with high-quality real multi-view data improves driving view reconstruction.

\begin{table}[t]
\caption{Ablation study on the design of GauBlend on VFHQ~\cite{xie2022vfhq}.}
\label{table:ablation_design}
\vspace{-6pt}
\centering
\resizebox{1\linewidth}{!}{
\begin{tabular}{l|ccc|ccc|cccc}
\toprule[1.2pt]
Method & \textit{D} & \textit{F} & \textit{R} & $\mathrm{B}$ & $reg$ & 2ST & PSNR $\uparrow$  & LPIPS $\downarrow$ & CSIM $\uparrow$ & AKD $\downarrow$ \\
\midrule
Baseline   & \checkmark & & & & & 
& 18.71 & 0.2880 & 0.8950 %0.5294 
& 2.9783\\
+$\mathrm{B}$        & \checkmark & & & \checkmark & & 
& 19.93 & 0.2449 & 0.9314 %0.7739 
& 2.6984 \\
+$\mathrm{B}$+$reg$     & \checkmark & & & \checkmark & \checkmark & 
& 20.10 & 0.2373 & 0.9344 %0.8004 
& 2.6692 \\
Variant-\textit{F}  & & \checkmark & & \checkmark & \checkmark & 
& 19.66 & 0.2548 & 0.9243 %0.7517 
& 2.8442 \\
Variant-\textit{R}  & & & \checkmark & \checkmark & \checkmark & 
& 19.71 & 0.2456 & 0.9289 %0.7683 
& 2.7609 \\
Ours & \textbf{\checkmark} & & & \textbf{\checkmark} & \textbf{\checkmark} & \textbf{\checkmark} 
& \textbf{20.41} & \textbf{0.2213} & \textbf{0.9373} %\textbf{0.8026} 
& \textbf{2.5571} \\
\bottomrule[1.2pt]
\end{tabular}
}

\vspace{2pt}
\footnotesize
\textit{\textit{D} = diffusion training; \textit{F} = fixed maximum-noise training; \textit{R} = regression; $\mathrm{B}$ = dynamic Gaussian blending bases; $reg$ = pixel-aligned Gaussian regularization; 2ST = two-stage training.}

\vspace{-8pt}
\end{table}

% \begin{table}[tb]
%   \centering
%   \resizebox{1\linewidth}{!}{
%   \begin{tabular}{l|cccccc} %cc} 
%   \toprule[1.2pt]
%   % \multirow{2}{*}{Method}  &  \multicolumn{7}{c|}{HDTF} \\
%   \multirow{2}{*}{Method} 
%   & 
%   & Sampling 
%   % Method
%              & %\multirow{2}{*}
%              {PSNR $\uparrow$}  & %\multirow{2}{*}
%              {SSIM $\uparrow$}   & %\multirow{2}{*}
%              {LPIPS $\downarrow$} & %\multirow{2}{*}
%              {CSIM $\uparrow$}  & %\multirow{2}{*}
%              {AKD $\downarrow$} %& AED $\downarrow$ & APD $\downarrow$
%              \\
%              & Steps            \\
%     \midrule
%     Diffusion & 1 
%     &  \\
%      + $\mathrm{Blend}$ & 1
%     &\\
%     + $\mathrm{Blend}$ + $reg$ & 1
%     &\\
%     Regression + $\mathrm{Blend}$ + $reg$ & 1
%     &\\
%     Non-Diffusion + $\mathrm{Blend}$ + $reg$ & 1
%     &\\

%     Diffusion + $\mathrm{Blend}$ + $reg$ + 2-stage & 30
%     &\\
%     Diffusion + $\mathrm{Blend}$ + $reg$ + 2-stage & 5
%     &\\
%     Diffusion + $\mathrm{Blend}$ + $reg$ + 2-stage & 2
%     &\\
%     Diffusion + $\mathrm{Blend}$ + $reg$ + 2-stage & 1
%     &\\
    
%     % \midrule
%     Diffusion + $\mathrm{Blend}$ + $reg$ + 2-stage (Ours) 
%     & 20.41 & 0.7988 & 0.2213 & 0.8026 & 2.5571
%     \\

%     \bottomrule[1.2pt]
%   \end{tabular}
%   }
%   \vspace{-8pt}
%   \caption{Ablation study on the design of our method. We present the results for self-reenactment on VFHQ~\cite{xie2022vfhq}.  
%   }
%   \label{table:ablation}

  % \vspace{-6pt}
% \end{table}
% \vspace{-3pt}
\section{Conclusion}
% \vspace{-5pt}

In this paper, we have presented GauBlend, a one-step diffusion model that directly generates canonical Gaussian avatars with dynamic Gaussian blending bases for one-shot animatable head avatar reconstruction. %The multi-timestep denoising of the diffusion model during training enables stable learning to denoise images of the subject with novel views and expressions to enhance direct canonical Gaussian avatar generation at each timestep, utilizing the additional appearance and motion cues to improve the reconstruction quality. 
GauBlend integrates a 3D canonical Gaussian avatar representation into a 2D diffusion pipeline, generating, animating and rendering it at each timestep during training to enable effective 2D supervision while preserving 3D consistency. 
The blending bases model diverse expression-dependent appearances and can be efficiently blended with driving expressions, enabling fast and expressive dynamic appearance modeling. Extensive experiments demonstrate high-quality avatar reconstruction and natural expression changes at 114 FPS.

\newpage
% llm statement
% \section{LLM Usage Statement}
% We used ChatGPT and OpenAI Codex to assist with language editing, grammatical refinement, related work organization, code refinement, and research idea refinement. These tools were also used to generate experimental comparison and data processing scripts, which were reviewed, verified and edited by the authors before use. No generative AI tools were used to produce figures, generate research data, or independently formulate scientific claims or conclusions. The authors retained full responsibility for the research methodology, experimental results, and final manuscript. The full manuscript or major portions thereof were not submitted to any LLM provider for reviewer-style feedback.

{    
    \small
    \bibliographystyle{ieeenat_fullname}
    \bibliography{main}

@String(PAMI = {IEEE Trans. Pattern Anal. Mach. Intell.})

@String(CVPR= {IEEE Conf. Comput. Vis. Pattern Recog.})

@String(ICCV= {Int. Conf. Comput. Vis.})

@String(ECCV= {Eur. Conf. Comput. Vis.})

@String(NIPS= {Adv. Neural Inform. Process. Syst.})

@String(TOG= {ACM Trans. Graph.})

@String(TVCG  = {IEEE Trans. Vis. Comput. Graph.})

@String(ICLR = {Int. Conf. Learn. Represent.})

@String(CVPRW= {IEEE Conf. Comput. Vis. Pattern Recog. Worksh.})

@String(PAMI  = {IEEE TPAMI})

@String(CVPR  = {CVPR})

@String(ICCV  = {ICCV})

@String(ECCV  = {ECCV})

@String(NIPS  = {NeurIPS})

@String(TOG   = {ACM TOG})

@String(TVCG  = {IEEE TVCG})

@String(ICLR  = {ICLR})

@String(CVPRW= {CVPRW})

@inproceedings{taubner2025cap4d,
  title={Cap4d: Creating animatable 4d portrait avatars with morphable multi-view diffusion models},
  author={Taubner, Felix and Zhang, Ruihang and Tuli, Mathieu and Lindell, David B},
  booktitle=CVPR,
  year={2025},
}

@inproceedings{taubner2025mvp4d,
  title={Mvp4d: Multi-view portrait video diffusion for animatable 4d avatars},
  author={Taubner, Felix and Zhang, Ruihang and Tuli, Mathieu and Bahmani, Sherwin and Lindell, David B},
  booktitle={SIGGRAPH Asia},
  year={2025}
}

@inproceedings{yin2025facecraft4d,
  title={Facecraft4d: Animated 3d facial avatar generation from a single image},
  author={Yin, Fei and Yao, Chun-Han and Mantiuk, Rafal K and Jampani, Varun and others},
  booktitle=ICCV,
  year={2025}
}

@inproceedings{wu2026uika,
  title={Uika: Fast universal head avatar from pose-free images},
  author={Wu, Zijian and Zhou, Boyao and Hu, Liangxiao and Liu, Hongyu and Sun, Yuan and Wang, Xuan and Cao, Xun and Shen, Yujun and Zhu, Hao},
  booktitle = CVPR,
  year      = {2026},
}

@inproceedings{mi2026one4d,
  title={One4d: Unified 4d generation and reconstruction via decoupled lora control},
  author={Mi, Zhenxing and Wang, Yuxin and Xu, Dan},
  booktitle={ECCV},
  year={2026}
}

@inproceedings{wang2026flashvggt,
  title={Flashvggt: Efficient and scalable visual geometry transformers with compressed descriptor attention},
  author={Wang, Zipeng and Xu, Dan},
  booktitle={CVPR},
  year={2026}
}

@inproceedings{he2025lam,
  title={LAM: Large Avatar Model for One-shot Animatable Gaussian Head},
  author={He, Yisheng and Gu, Xiaodong and Ye, Xiaodan and Xu, Chao and Zhao, Zhengyi and Dong, Yuan and Yuan, Weihao and Dong, Zilong and Bo, Liefeng},
  booktitle={SIGGRAPH},
  year={2025}
}

@inproceedings{zhao2026one,
  title={One-Shot Feed-Forward 360$^\circ$ Animatable Avatar via Inpainted UV-Space Gaussian Modeling},
  author={Zhao, Shuling and Xu, Dan},
  booktitle=ECCV,
  year={2026},
}

@inproceedings{kirschstein2026flexavatar,
  title={Flexavatar: Learning complete 3d head avatars with partial supervision},
  author={Kirschstein, Tobias and Giebenhain, Simon and Nie{\ss}ner, Matthias},
  booktitle=CVPR,
  year={2026}
}

@article{kerbl20233d,
  title={3D Gaussian splatting for real-time radiance field rendering.},
  author={Kerbl, Bernhard and Kopanas, Georgios and Leimk{\"u}hler, Thomas and Drettakis, George},
  journal=TOG, 
  volume={42},
  number={4},
  pages={139--1},
  year={2023}
}

@inproceedings{cai2025baking,
  title={Baking gaussian splatting into diffusion denoiser for fast and scalable single-stage image-to-3d generation and reconstruction},
  author={Cai, Yuanhao and Zhang, He and Zhang, Kai and Liang, Yixun and Ren, Mengwei and Luan, Fujun and Liu, Qing and Kim, Soo Ye and Zhang, Jianming and Zhang, Zhifei and others},
  booktitle=ICCV,
  year={2025},
}

@inproceedings{ma20243d,
  title={3d gaussian blendshapes for head avatar animation},
  author={Ma, Shengjie and Weng, Yanlin and Shao, Tianjia and Zhou, Kun},
  booktitle={SIGGRAPH},
  year={2024}
}

@inproceedings{dhamo2024headgas,
  title={Headgas: Real-time animatable head avatars via 3d gaussian splatting},
  author={Dhamo, Helisa and Nie, Yinyu and Moreau, Arthur and Song, Jifei and Shaw, Richard and Zhou, Yiren and P{\'e}rez-Pellitero, Eduardo},
  booktitle=ECCV,
  year={2024},
}

@inproceedings{li2025rgbavatar,
  title={Rgbavatar: Reduced gaussian blendshapes for online modeling of head avatars},
  author={Li, Linzhou and Li, Yumeng and Weng, Yanlin and Zheng, Youyi and Zhou, Kun},
  booktitle=CVPR,
  year={2025},
}

@inproceedings{liu2026avatarpointillist,
    author    = {Liu, Hongyu and Wang, Xuan and Wu, Zijian and Wang, Yating and Wan, Ziyu and Ma, Yue and Liu, Runtao and Zhou, Boyao and Shen, Yujun and Chen, Qifeng},
    title     = {AvatarPointillist: AutoRegressive 4D Gaussian Avatarization},
    booktitle = CVPR,
    year      = {2026},
}

@inproceedings{peng2026flexavatar,
  title={FlexAvatar: flexible large reconstruction model for animatable gaussian head avatars with detailed deformation},
  author={Peng, Cheng and Su, Zhuo and Wang, Liao and Guo, Chen and Li, Zhaohu and Long, Chengjiang and Lv, Zheng and Sun, Jingxiang and Zhang, Chenyangguang and Liu, Yebin},
  booktitle=CVPR,
  year={2026}
}

@inproceedings{zhou2025zero,
  title={Zero-1-to-A: Zero-Shot One Image to Animatable Head Avatars Using Video Diffusion},
  author={Zhou, Zhenglin and Ma, Fan and Fan, Hehe and Chua, Tat-Seng},
  booktitle=CVPR,
  year={2025}
}

@inproceedings{ma2024follow,
  title={Follow-your-emoji: Fine-controllable and expressive freestyle portrait animation},
  author={Ma, Yue and Liu, Hongyu and Wang, Hongfa and Pan, Heng and He, Yingqing and Yuan, Junkun and Zeng, Ailing and Cai, Chengfei and Shum, Heung-Yeung and Liu, Wei and others},
  booktitle={SIGGRAPH Asia},
  year={2024}
}

@inproceedings{rombach2022high,
  title={High-resolution image synthesis with latent diffusion models},
  author={Rombach, Robin and Blattmann, Andreas and Lorenz, Dominik and Esser, Patrick and Ommer, Bj{\"o}rn},
  booktitle=CVPR,
  year={2022}
}

@article{guo2024liveportrait,
  title={Liveportrait: Efficient portrait animation with stitching and retargeting control},
  author={Guo, Jianzhu and Zhang, Dingyun and Liu, Xiaoqiang and Zhong, Zhizhou and Zhang, Yuan and Wan, Pengfei and Zhang, Di},
  journal={arXiv preprint arXiv:2407.03168},
  year={2024}
}

@inproceedings{khakhulin2022realistic,
  title={Realistic one-shot mesh-based head avatars},
  author={Khakhulin, Taras and Sklyarova, Vanessa and Lempitsky, Victor and Zakharov, Egor},
  booktitle=ECCV,
  year={2022},
}

@inproceedings{ye2024realdportrait,
title={Real3D-Portrait: One-shot Realistic 3D Talking Portrait Synthesis},
author={Zhenhui Ye and Tianyun Zhong and Yi Ren and Jiaqi Yang and Weichuang Li and Jiawei Huang and Ziyue Jiang and Jinzheng He and Rongjie Huang and Jinglin Liu and Chen Zhang and Xiang Yin and Zejun MA and Zhou Zhao},
booktitle=ICLR,
year={2024},
}

@inproceedings{deng2024portrait4d,
  title={c},
  author={Deng, Yu and Wang, Duomin and Ren, Xiaohang and Chen, Xingyu and Wang, Baoyuan},
  booktitle=CVPR,
  year={2024}
}

@inproceedings{deng2024portrait4d_eccv,
  title={Portrait4d-v2: Pseudo multi-view data creates better 4d head synthesizer},
  author={Deng, Yu and Wang, Duomin and Wang, Baoyuan},
  booktitle=ECCV,
  year={2024},
}

@inproceedings{chu2024generalizable,
  title={Generalizable and animatable gaussian head avatar},
  author={Chu, Xuangeng and Harada, Tatsuya},
  booktitle=NIPS,
  year={2024}
}

@inproceedings{li2023generalizable,
  title={Generalizable one-shot 3d neural head avatar},
  author={Li, Xueting and De Mello, Shalini and Liu, Sifei and Nagano, Koki and Iqbal, Umar and Kautz, Jan},
  booktitle=NIPS,
  year={2023}
}

@inproceedings{zhu2022celebv,
  title={CelebV-HQ: A large-scale video facial attributes dataset},
  author={Zhu, Hao and Wu, Wayne and Zhu, Wentao and Jiang, Liming and Tang, Siwei and Zhang, Li and Liu, Ziwei and Loy, Chen Change},
  booktitle=ECCV,
  year={2022},
}

@inproceedings{chung18b_interspeech,
  title     = {{VoxCeleb2: Deep Speaker Recognition}},
  author    = {Joon Son Chung and Arsha Nagrani and Andrew Zisserman},
  year      = {2018},
  booktitle = {{Interspeech}},
}

@inproceedings{xie2022vfhq,
  title={Vfhq: A high-quality dataset and benchmark for video face super-resolution},
  author={Xie, Liangbin and Wang, Xintao and Zhang, Honglun and Dong, Chao and Shan, Ying},
  booktitle=CVPRW,
  year={2022}
}

@inproceedings{karras2019style,
  title={A style-based generator architecture for generative adversarial networks},
  author={Karras, Tero and Laine, Samuli and Aila, Timo},
  booktitle=cvpr,
  year={2019}
}

@InProceedings{ren2026omg,
    author    = {Ren, Jianqiang and Liu, Lin and Hoi, Steven},
    title     = {OMG-Avatar: One-shot Multi-LOD Gaussian Head Avatar},
    booktitle = CVPR,
    year      = {2026},
}

@inproceedings{yao2026ffavatar,
  title={FFAvatar: Feed-Forward 4D Head Avatar Reconstruction from Sparse Portrait Images},
  author={Yao, Jianjiang and Xian, Ke and Dai, Renxiang and Qiu, Robert Caiming},
  booktitle=ECCV,
  year={2026},
}

@article{li2017learning,
  title={Learning a model of facial shape and expression from 4D scans},
  author={Li, Tianye and Bolkart, Timo and Black, Michael J and Li, Hao and Romero, Javier},
  journal=TOG, 
  volume={36},
  number={6},
  pages={1--17},
  year={2017},
  publisher={Association for Computing Machinery (ACM)}
}

@article{cao2013facewarehouse,
  title={Facewarehouse: A 3d facial expression database for visual computing},
  author={Cao, Chen and Weng, Yanlin and Zhou, Shun and Tong, Yiying and Zhou, Kun},
  journal=TVCG, 
  volume={20},
  number={3},
  pages={413--425},
  year={2013},
  publisher={IEEE}
}

@inproceedings{blanz1999morphable,
  title={A Morphable Model for the Synthesis of 3D Faces},
  author={Blanz, Volker and Vetter, Thomas},
  booktitle={SIGGRAPH},
  year={1999},
}

@inproceedings{paysan20093d,
  title={A 3D face model for pose and illumination invariant face recognition},
  author={Paysan, Pascal and Knothe, Reinhard and Amberg, Brian and Romdhani, Sami and Vetter, Thomas},
  booktitle={International Conference on Advanced Video and Signal Based Surveillance},
  year={2009},
}

@inproceedings{vlasic2006face,
  title={Face transfer with multilinear models},
  author={Vlasic, Daniel and Brand, Matthew and Pfister, Hanspeter and Popovic, Jovan},
  booktitle={ACM SIGGRAPH 2006 Courses},
  year={2006}
}

@article{ploumpis2020towards,
  title={Towards a complete 3D morphable model of the human head},
  author={Ploumpis, Stylianos and Ververas, Evangelos and O'Sullivan, Eimear and Moschoglou, Stylianos and Wang, Haoyang and Pears, Nick and Smith, William AP and Gecer, Baris and Zafeiriou, Stefanos},
  journal=PAMI,
  volume={43},
  number={11},
  pages={4142--4160},
  year={2020},
  publisher={IEEE}
}

@inproceedings{hong2022headnerf,
  title={HeadNeRF: A realtime NeRF-based parametric head model},
  author={Hong, Yang and Peng, Bo and Xiao, Haiyao and Liu, Ligang and Zhang, Juyong},
  booktitle=CVPR,
  year={2022},
}

@inproceedings{yenamandra2021i3dmm,
  title={i3dmm: Deep implicit 3d morphable model of human heads},
  author={Yenamandra, Tarun and Tewari, Ayush and Bernard, Florian and Seidel, Hans-Peter and Elgharib, Mohamed and Cremers, Daniel and Theobalt, Christian},
  booktitle=CVPR,
  year={2021},
}

@inproceedings{giebenhain2023learning,
  title={Learning neural parametric head models},
  author={Giebenhain, Simon and Kirschstein, Tobias and Georgopoulos, Markos and R{\"u}nz, Martin and Agapito, Lourdes and Nie{\ss}ner, Matthias},
  booktitle=CVPR,
  year={2023},
}

@article{gao2022reconstructing,
  title={Reconstructing personalized semantic facial nerf models from monocular video},
  author={Gao, Xuan and Zhong, Chenglai and Xiang, Jun and Hong, Yang and Guo, Yudong and Zhang, Juyong},
  journal=TOG,
  volume={41},
  number={6},
  pages={1--12},
  year={2022},
  publisher={ACM, New York, NY, USA}
}

@inproceedings{anciukevivcius2023renderdiffusion,
  title={Renderdiffusion: Image diffusion for 3d reconstruction, inpainting and generation},
  author={Anciukevi{\v{c}}ius, Titas and Xu, Zexiang and Fisher, Matthew and Henderson, Paul and Bilen, Hakan and Mitra, Niloy J and Guerrero, Paul},
  booktitle=CVPR,
  year={2023},
}

@article{ho2020denoising,
  title={Denoising diffusion probabilistic models},
  author={Ho, Jonathan and Jain, Ajay and Abbeel, Pieter},
  journal=NIPS,
  year={2020}
}

@inproceedings{xu2024dmv3d,
  title={Dmv3d: Denoising multi-view diffusion using 3d large reconstruction model},
  author={Xu, Yinghao and Tan, Hao and Luan, Fujun and Bi, Sai and Wang, Peng and Li, Jiahao and Shi, Zifan and Sunkavalli, Kalyan and Wetzstein, Gordon and Xu, Zexiang and others},
  booktitle=ICLR,
  year={2024}
}

@inproceedings{qian2024gaussianavatars,
  title={Gaussianavatars: Photorealistic head avatars with rigged 3d gaussians},
  author={Qian, Shenhan and Kirschstein, Tobias and Schoneveld, Liam and Davoli, Davide and Giebenhain, Simon and Nie{\ss}ner, Matthias},
  booktitle=CVPR,
  year={2024}
}

@inproceedings{szymanowicz2024splatter,
  title={Splatter image: Ultra-fast single-view 3d reconstruction},
  author={Szymanowicz, Stanislaw and Rupprecht, Chrisitian and Vedaldi, Andrea},
  booktitle=CVPR,
  year={2024}
}

@inproceedings{tang2024lgm,
  title={Lgm: Large multi-view gaussian model for high-resolution 3d content creation},
  author={Tang, Jiaxiang and Chen, Zhaoxi and Chen, Xiaokang and Wang, Tengfei and Zeng, Gang and Liu, Ziwei},
  booktitle=ECCV,
  year={2024},
}

@inproceedings{xu2024grm,
  title={Grm: Large gaussian reconstruction model for efficient 3d reconstruction and generation},
  author={Xu, Yinghao and Shi, Zifan and Yifan, Wang and Chen, Hansheng and Yang, Ceyuan and Peng, Sida and Shen, Yujun and Wetzstein, Gordon},
  booktitle=ECCV,
  year={2024},
}

@inproceedings{zhang2024gs,
  title={Gs-lrm: Large reconstruction model for 3d gaussian splatting},
  author={Zhang, Kai and Bi, Sai and Tan, Hao and Xiangli, Yuanbo and Zhao, Nanxuan and Sunkavalli, Kalyan and Xu, Zexiang},
  booktitle=ECCV,
  year={2024},
}

@inproceedings{wang2024pf,
  title={Pf-lrm: Pose-free large reconstruction model for joint pose and shape prediction},
  author={Wang, Peng and Tan, Hao and Bi, Sai and Xu, Yinghao and Luan, Fujun and Sunkavalli, Kalyan and Wang, Wenping and Xu, Zexiang and Zhang, Kai},
  booktitle=ICLR,
  year={2024}
}

@article{dumoulin2016guide,
  title={A guide to convolution arithmetic for deep learning},
  author={Dumoulin, Vincent and Visin, Francesco},
  journal={arXiv preprint arXiv:1603.07285},
  year={2016}
}

@inproceedings{kirschstein2024gghead,
  title={Gghead: Fast and generalizable 3d gaussian heads},
  author={Kirschstein, Tobias and Giebenhain, Simon and Tang, Jiapeng and Georgopoulos, Markos and Nie{\ss}ner, Matthias},
  booktitle={SIGGRAPH Asia},
  year={2024}
}

@inproceedings{an2023panohead,
  title={Panohead: Geometry-aware 3d full-head synthesis in 360deg},
  author={An, Sizhe and Xu, Hongyi and Shi, Yichun and Song, Guoxian and Ogras, Umit Y and Luo, Linjie},
  booktitle=CVPR,
  year={2023}
}

@inproceedings{bilecen2024dual,
  title={Dual encoder GAN inversion for high-fidelity 3D head reconstruction from single images},
  author={Bilecen, Bahri Batuhan and G{\"o}kmen, Ahmet and Dundar, Aysegul},
  booktitle=NIPS,
  year={2024}
}

@inproceedings{li2026large,
  title={Large-scale Codec Avatars: The Unreasonable Effectiveness of Large-scale Avatar Pretraining},
  author={Li, Junxuan and Khirodkar, Rawal and Zakharov, Egor and Lee, Jihyun and Su, Zhaoen and Dong, Yuan and Martinez, Julieta and Li, Kai and Tan, Qingyang and Shiratori, Takaaki and others},
  booktitle=CVPR,
  year={2026}
}

@inproceedings{hong2024lrm,
  title={Lrm: Large reconstruction model for single image to 3d},
  author={Hong, Yicong and Zhang, Kai and Gu, Jiuxiang and Bi, Sai and Zhou, Yang and Liu, Difan and Liu, Feng and Sunkavalli, Kalyan and Bui, Trung and Tan, Hao},
  booktitle=ICLR,
  year={2024}
}

@article{xie2024lrm,
  title={Lrm-zero: Training large reconstruction models with synthesized data},
  author={Xie, Desai and Bi, Sai and Shu, Zhixin and Zhang, Kai and Xu, Zexiang and Zhou, Yi and Pirk, S{\"o}ren and Kaufman, Arie and Sun, Xin and Tan, Hao},
  journal=NIPS,
  year={2024}
}

@inproceedings{zou2024triplane,
  title={Triplane meets gaussian splatting: Fast and generalizable single-view 3d reconstruction with transformers},
  author={Zou, Zi-Xin and Yu, Zhipeng and Guo, Yuan-Chen and Li, Yangguang and Liang, Ding and Cao, Yan-Pei and Zhang, Song-Hai},
  booktitle=CVPR,
  year={2024}
}

@inproceedings{ziwen2025long,
  title={Long-lrm: Long-sequence large reconstruction model for wide-coverage gaussian splats},
  author={Ziwen, Chen and Tan, Hao and Zhang, Kai and Bi, Sai and Luan, Fujun and Hong, Yicong and Fuxin, Li and Xu, Zexiang},
  booktitle=ICCV,
  year={2025},
}

@article{zhou2024diffgs,
  title={Diffgs: Functional gaussian splatting diffusion},
  author={Zhou, Junsheng and Zhang, Weiqi and Liu, Yu-Shen},
  journal=NIPS,
  year={2024}
}

@inproceedings{lin2025diffsplat,
  title={Diffsplat: Repurposing image diffusion models for scalable gaussian splat generation},
  author={Lin, Chenguo and Pan, Panwang and Yang, Bangbang and Li, Zeming and Mu, Yadong},
  booktitle=ICLR,
  year={2025}
}

@article{kirschstein2023nersemble,
  title={Nersemble: Multi-view radiance field reconstruction of human heads},
  author={Kirschstein, Tobias and Qian, Shenhan and Giebenhain, Simon and Walter, Tim and Nie{\ss}ner, Matthias},
  journal=TOG,
  volume={42},
  number={4},
  pages={1--14},
  year={2023},
  publisher={ACM, New York, NY, USA}
}

@article{martinez2024codec,
  title={Codec avatar studio: Paired human captures for complete, driveable, and generalizable avatars},
  author={Martinez, Julieta and Kim, Emily and Romero, Javier and Bagautdinov, Timur and Saito, Shunsuke and Yu, Shoou-I and Anderson, Stuart and Zollh{\"o}fer, Michael and Wang, Te-Li and Bai, Shaojie and others},
  journal=NIPS,
  year={2024}
}

@inproceedings{zhang2021flow,
  title={Flow-guided one-shot talking face generation with a high-resolution audio-visual dataset},
  author={Zhang, Zhimeng and Li, Lincheng and Ding, Yu and Fan, Changjie},
  booktitle=CVPR,
  year={2021}
}

@inproceedings{zhang2018unreasonable,
  title={The unreasonable effectiveness of deep features as a perceptual metric},
  author={Zhang, Richard and Isola, Phillip and Efros, Alexei A and Shechtman, Eli and Wang, Oliver},
  booktitle=CVPR,
  year={2018}
}

@inproceedings{bulat2017far,
  title={How far are we from solving the 2d \& 3d face alignment problem?(and a dataset of 230,000 3d facial landmarks)},
  author={Bulat, Adrian and Tzimiropoulos, Georgios},
  booktitle=ICCV,
  year={2017}
}

@inproceedings{retsinas20243d,
  title={3d facial expressions through analysis-by-neural-synthesis},
  author={Retsinas, George and Filntisis, Panagiotis P and Danecek, Radek and Abrevaya, Victoria F and Roussos, Anastasios and Bolkart, Timo and Maragos, Petros},
  booktitle=CVPR,
  year={2024}
}

@article{lugaresi2019mediapipe,
  title={Mediapipe: A framework for building perception pipelines},
  author={Lugaresi, Camillo and Tang, Jiuqiang and Nash, Hadon and McClanahan, Chris and Uboweja, Esha and Hays, Michael and Zhang, Fan and Chang, Chuo-Ling and Yong, Ming Guang and Lee, Juhyun and others},
  journal={arXiv preprint arXiv:1906.08172},
  year={2019}
}

@inproceedings{song2020denoising,
  title={Denoising diffusion implicit models},
  author={Song, Jiaming and Meng, Chenlin and Ermon, Stefano},
  booktitle =ICLR,
  year={2021}
}

@article{amos2016openface,
  title={Openface: A general-purpose face recognition library with mobile applications},
  author={Amos, Brandon and Ludwiczuk, Bartosz and Satyanarayanan, Mahadev and others},
  journal={CMU School of Computer Science},
  volume={6},
  number={2},
  pages={20},
  year={2016},
  publisher={CMU School of Computer Science, Pittsburgh, PA}
}

@inproceedings{peebles2023scalable,
  title={Scalable diffusion models with transformers},
  author={Peebles, William and Xie, Saining},
  booktitle=ICCV,
  year={2023},
}

@article{harkonen2020ganspace,
  title={Ganspace: Discovering interpretable gan controls},
  author={H{\"a}rk{\"o}nen, Erik and Hertzmann, Aaron and Lehtinen, Jaakko and Paris, Sylvain},
  journal=NIPS,
  year={2020}
}

@inproceedings{nichol2021improved,
  title={Improved denoising diffusion probabilistic models},
  author={Nichol, Alexander Quinn and Dhariwal, Prafulla},
  booktitle={ICML},
  year={2021},
}

@inproceedings{loshchilov2017decoupled,
  author       = {Loshchilov, Ilya and Hutter, Frank},
  title        = {Decoupled Weight Decay Regularization},
  booktitle    = ICLR,
  year         = {2019},
}

@article{chicherin2023adversarially,
  title={Adversarially-guided portrait matting},
  author={Chicherin, Sergej and Efremyan, Karen},
  journal={arXiv preprint arXiv:2305.02981},
  year={2023}
}

@inproceedings{liu2026efficient,
  title={Efficient and High-Quality Depth Estimation via Pixel-Space Diffusion with Linear Attention},
  author={Liu, Bingde and Ran, Wu and Zhang, Jinglei and Yuan, Huanhuan and Ma, Chao},
  booktitle=ECCV,
  year={2026},
}
}

\clearpage
\setcounter{page}{1}
\maketitlesupplementary

\appendix

In this supplementary material, we provide the implementation details in Sec.~\ref{sec:implement}, additional experimental results in Sec.~\ref{sec:supp_exp}, and discuss the limitations in Sec.~\ref{sec:limitations}.

\section{Implementation Details}
\label{sec:implement}
We use the diffusion transformer~\cite{peebles2023scalable} with $L=14$ layers as our backbone and adopt the cosine noise schedule~\cite{nichol2021improved} with 1000 diffusion timesteps.
The model takes $N=2$ noisy input images. We represent the canonical Gaussian avatar in a UV space of size $K=256$. 
As the input expression code to our dynamic Gaussian basis blending module, we use the FLAME~\cite{li2017learning} expression code with dimension $d_e=100$. 
The generated dynamic Gaussian blending basis consists of $T=15$ blending offsets, each with dimension $d_b=32$.

We train the model using the AdamW optimizer~\cite{loshchilov2017decoupled} with a learning rate of $5\times 10^{-4}$. For supervision, we use one real novel-view image. When incorporating pseudo-multi-view supervision from the pretrained 3D GAN inversion model~\cite{an2023panohead, bilecen2024dual}, we additionally sample one pseudo-view uniformly from the camera azimuth range $[0,360^\circ)$ at an elevation of $0^\circ$. Thus, the number of novel views is $M=2$ during the first training stage and $M=1$ during the second stage. We use a batch size of 16 and train for 90K steps in the first stage and 10K steps in the second stage on an NVIDIA H20 GPU. In the second stage, we follow LCA~\cite{li2026large} and use differential learning rates, with a layer-wise learning-rate decay factor of $0.65$. The loss weights are set to $\lambda=0.5$ and $\lambda^{\alpha}=1$. For the first 12K steps, we set the loss weight $\beta$ in $\mathcal{L}_{reg}^{\mu}$ to 0 to prevent the unstable pixel-aligned Gaussian position maps from affecting the prediction of the canonical Gaussian positions; thereafter, we set $\beta=0.25$. We introduce the identity similarity loss $\mathcal{L}_{id}$ in $\mathcal{L}_{rec}$ after 20K steps. Finally, we set the threshold $\epsilon$ in $\mathcal{L}_{\Delta\mu}$ to $0.1$.

\section{Additional Experimental Results}
\label{sec:supp_exp}

\subsection{Additional Evaluation Details}
\label{sec:supp_eval_details}
We compare our method with several recent state-of-the-art approaches, including Portrait4D-v2~\cite{deng2024portrait4d_eccv}, GAGAvatar~\cite{chu2024generalizable}, LAM~\cite{he2025lam}, UIKA~\cite{wu2026uika}, and FlexAvatar~\cite{kirschstein2026flexavatar}. For each method, we preprocess the data using the official procedure, and use the released model checkpoints and animation scripts to generate the final results. To compute the quantitative metrics, we set the backgrounds of both the ground-truth and generated frames to white. Since GAGAvatar only generates frames with black backgrounds, we use StyleMatte~\cite{chicherin2023adversarially} to extract the foreground subjects and replace the backgrounds with white before computing the metrics. Because Portrait4D-v2 does not produce spatially aligned animation results with the driving images, we align its output before computing the self-reenactment metrics.

% \subsection{Additional Experimental Results}

\subsection{Video Results}
% We provide an offline web page in the supplementary material with video results, 
We provide video results on the project page\footnote{\href{https://shaelynz.github.io/gaublend_project/}{https://shaelynz.github.io/gaublend\_project/}}, 
including comparisons with state-of-the-art methods and ablation studies. To avoid the temporal inconsistencies caused by the foreground extraction and alignment procedures for GAGAvatar~\cite{chu2024generalizable} and Portrait4D-v2~\cite{deng2024portrait4d_eccv} described in Sec.~\ref{sec:supp_eval_details}, we omit these procedures when generating the video comparison results.
By blending the dynamic Gaussian bases to model expression-dependent appearance, our method produces videos with natural dynamic appearance changes, such as wrinkles appearing and disappearing. Visualizations of the generated Gaussian blending bases demonstrate their ability to capture diverse dynamic appearances, while visualizations of the blending weights during animation show how they vary with the expression code to blend the bases and produce the target expression-dependent appearance.
% We further present more visualization results for the learned blending bases.

\begin{figure*}[tb]
    \centering
  \includegraphics[width=0.99\linewidth]{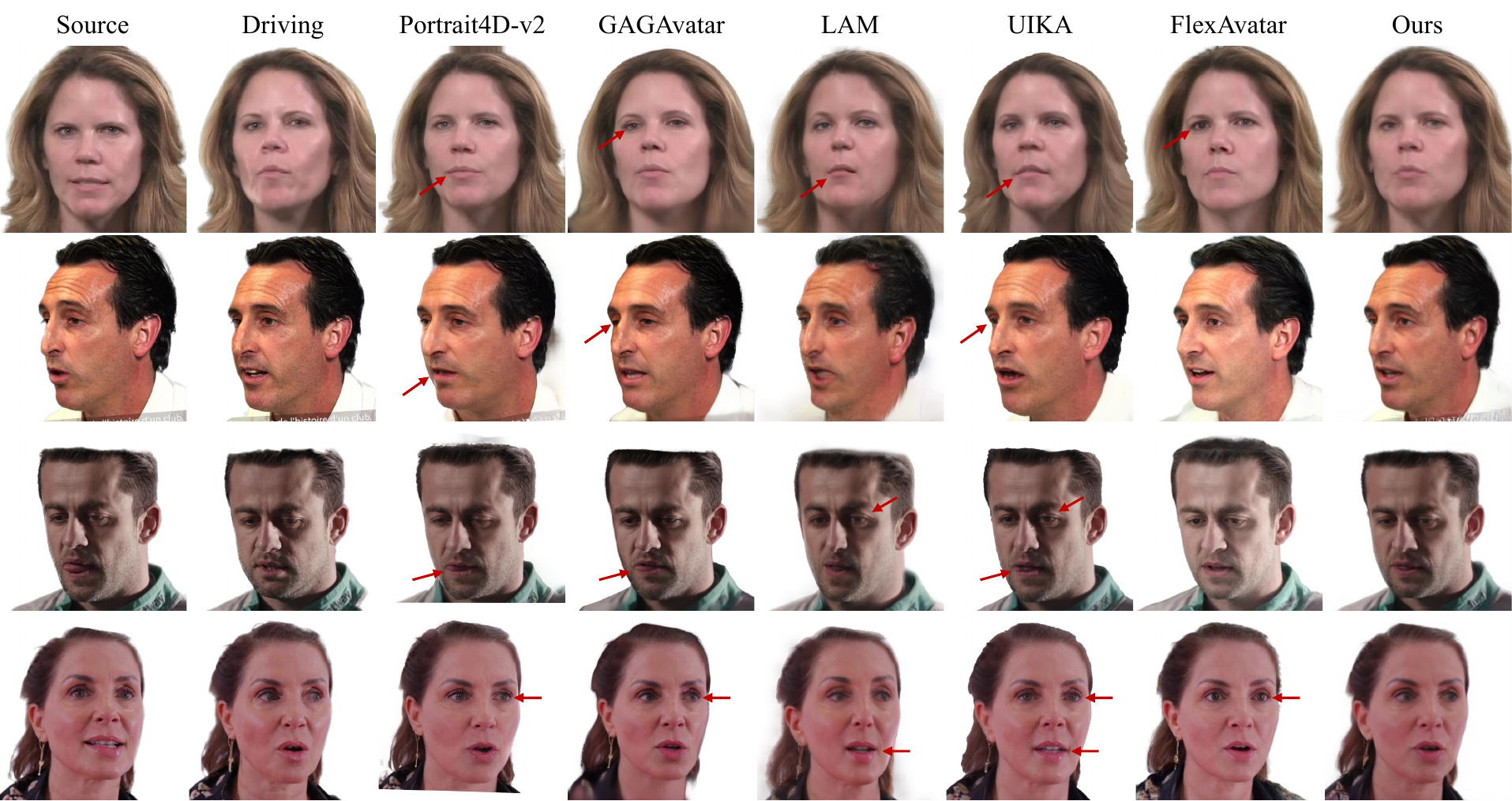}

  \vspace{-9pt}
  \caption{Additional qualitative comparison with state-of-the-art methods for self-reenactment on the VFHQ~\cite{xie2022vfhq} and HDTF~\cite{zhang2021flow} datasets. Our method follows the driving motion accurately while preserving source identity.
  }
  % \vspace{-4pt}
  \label{fig:sota_cmp_sameid_supp}
\end{figure*}

\begin{figure*}[tb]
    \centering
  \includegraphics[width=0.99\linewidth]{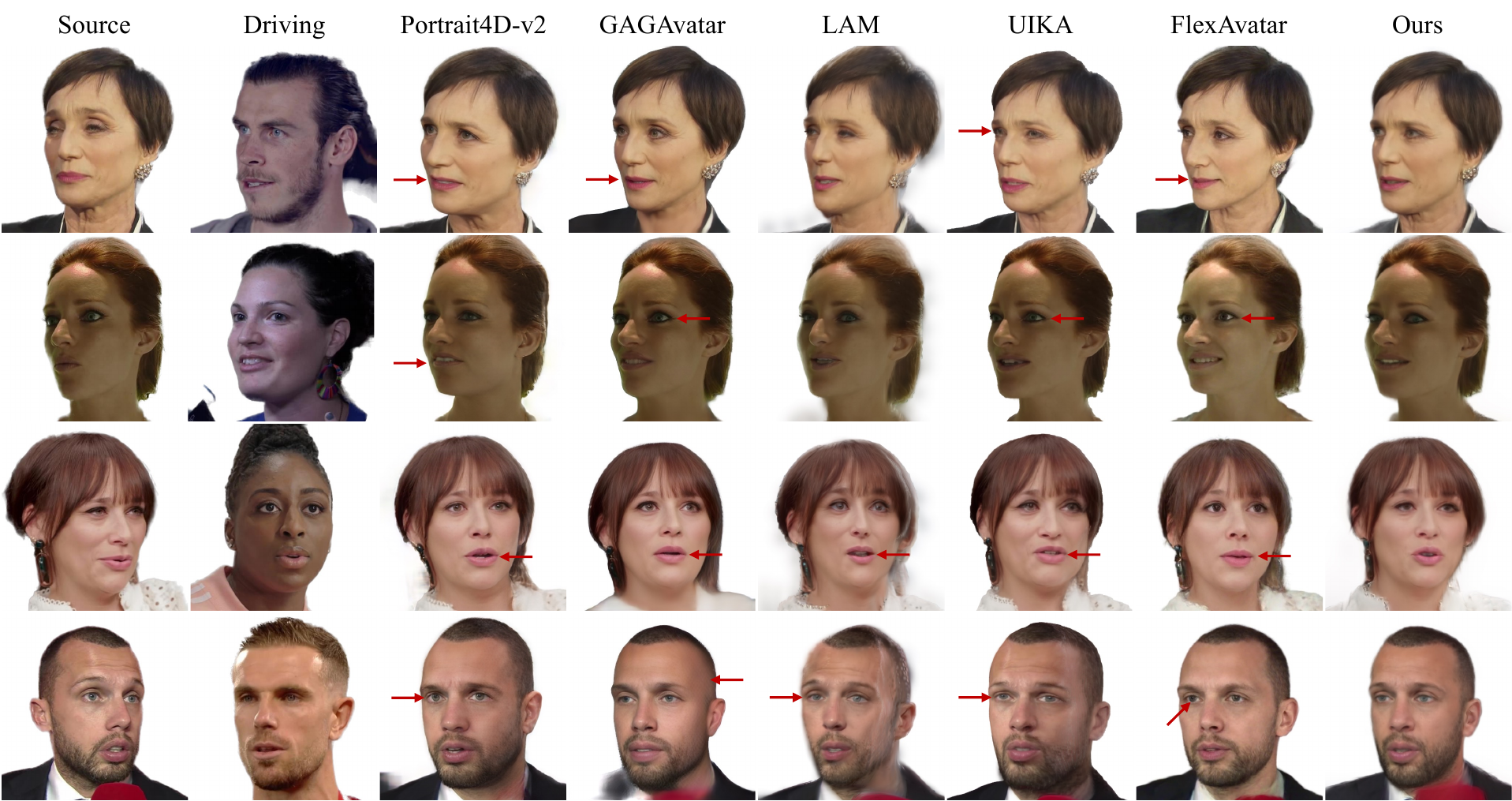}

  \vspace{-9pt}
  \caption{Additional qualitative comparison with state-of-the-art methods for cross-identity reenactment on the VFHQ~\cite{xie2022vfhq} dataset. Our method follows the driving motion accurately while preserving source identity.
  }
  % \vspace{-4pt}
  \label{fig:sota_cmp_crossid_supp}
\end{figure*}

\subsection{Additional Comparison Results}

\noindent\textbf{Additional Qualitative Comparison Results.}
We provide additional qualitative comparison results for self-reenactment in Fig.~\ref{fig:sota_cmp_sameid_supp} and cross-identity reenactment in Fig.~\ref{fig:sota_cmp_crossid_supp}. Our method best captures the driving motion, particularly around the eyes and mouth, where facial movements are most dynamic. For example, it accurately reproduces pouting expressions (see row 1 in Fig.~\ref{fig:sota_cmp_sameid_supp} and row 3 in Fig.~\ref{fig:sota_cmp_crossid_supp}). This demonstrates the expressiveness of the generated dynamic Gaussian blending bases for modeling locally varying appearances. Regarding source-identity preservation, FlexAvatar~\cite{kirschstein2026flexavatar} struggles to preserve the source skin tone (see rows 2 and 3 in Fig.~\ref{fig:sota_cmp_sameid_supp}) and eye color (see rows 2 and 4 in Fig.~\ref{fig:sota_cmp_crossid_supp}) during feed-forward inference. In contrast, our method effectively preserves the source appearance.

\begin{table}[tb]
\caption{User preference rates for our method versus other state-of-the-art methods.}
\vspace{-18pt}
  \begin{center}
  \resizebox{1\linewidth}{!}{
  \begin{tabular}{l|cccc}
  \toprule[1.2pt]
     {Comparison} & { vs. Portrait4D-v2~\cite{deng2024portrait4d_eccv}} & { vs. GAGAvatar~\cite{chu2024generalizable} }& { vs. UIKA~\cite{wu2026uika}} & { vs. FlexAvatar~\cite{kirschstein2026flexavatar}} \\
    \midrule
     {Ours Preferred $\uparrow$} &{ 89.5\%} &{ 85.9\%} &{ 93.6\%} &{ 80.5\%}   \\  
    \bottomrule[1.2pt]
  \end{tabular}
  }
  \end{center}
  % \vspace{-22pt}
  
  % \vspace{-13pt}
  \label{supp:table:user_study}
\end{table}

\noindent\textbf{User Study for Cross-identity Reenactment.}
Since ground-truth data are unavailable for cross-identity reenactment, we conduct a user study with 22 participants, who compare 10 cross-identity reenactment results produced by our method and other state-of-the-art approaches. We exclude LAM~\cite{he2025lam} from the comparison because it achieves the lowest quantitative performance. The participants are asked to evaluate the results based on source-identity preservation, driving-motion transfer, and overall quality. As shown in Tab.~\ref{supp:table:user_study}, participants prefer our results, confirming the high quality of our animations. In particular, our method is strongly preferred over UIKA~\cite{wu2026uika}, which does not model dynamic appearance. This highlights the importance of expression-dependent appearance modeling for natural avatar animation.

\subsection{Additional Ablation Study}

\begin{figure}[tb]
    \centering
  \includegraphics[width=1\linewidth]{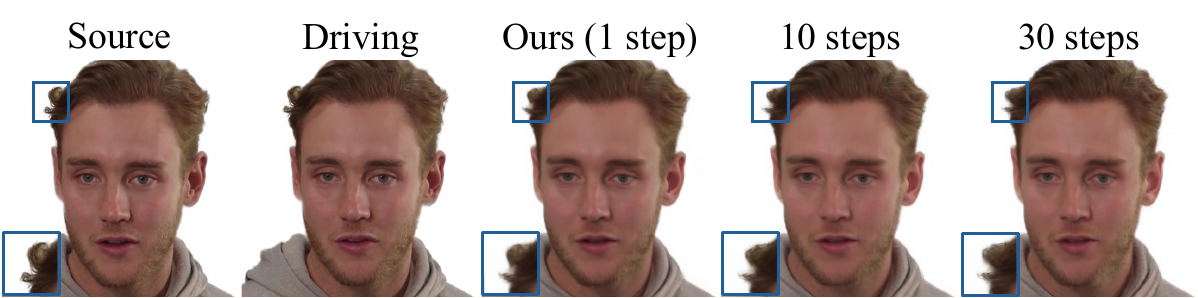}

  \vspace{-9pt}
  \caption{Qualitative ablation study on the number of sampling steps during inference.
  }
  % \vspace{-4pt}
  \label{fig:abla_samplingsteps_supp}
\end{figure}

\noindent\textbf{Number of Sampling Steps.}
As shown in Tab.~\ref{table:ablation_sampling_steps}, one-step sampling achieves the best performance during inference. Although multi-step DDIM sampling~\cite{song2020denoising} produces more natural denoised views by adapting the canonical avatar to novel viewpoints and motions, it reduces fidelity to the source image, particularly in local details, as shown in Fig.~\ref{fig:abla_samplingsteps_supp}. This observation is consistent with findings in other domains~\cite{liu2026efficient}. We therefore use one-step inference to achieve both high quality and efficiency.

%will make the generated canonical avatar adapt to the denoised novel views and motion. Although it can make the denoised images more natural, it becomes less faithful to the source image, losing appearance details from the source image. Therefore, we adopt one-step inference, which achieves both efficiency and high quality.

\begin{table}[tb]
\caption{Ablation study on the number of sampling steps during inference on VFHQ~\cite{xie2022vfhq}.
  }
  \label{table:ablation_sampling_steps}

  \vspace{-6pt}
  \centering
  \resizebox{1\linewidth}{!}{
  \begin{tabular}{l|ccccccc} 
  \toprule[1.2pt]
  {\#Steps}  
             & PSNR $\uparrow$  & SSIM $\uparrow$   & LPIPS $\downarrow$ & CSIM $\uparrow$  & AKD $\downarrow$ & AED $\downarrow$ & APD $\downarrow$
             \\
    \midrule
    30 
    & 19.28 & 0.7841 & 0.2339 & 0.9304 %0.7662 
    & 2.5984 & 0.3302
    & 0.8970
    \\
    10 
    & 19.52 & 0.7908 & 0.2272 & 0.9339 %0.7871 
    & 2.5964 & 0.3298 & 0.8580
    \\
    5
    & 19.68 & 0.7921 & 0.2251 & 0.9353 %0.7900 
    & 2.5739 & 0.3265 & 0.8450
    \\
    2 
    & 19.93 & 0.7935 & 0.2240 & 0.9342 %0.7993 
    & 2.5955 & 0.3297 & 0.8488
    \\
    
    % \midrule
    1 (Ours) 
    & \textbf{20.41} & \textbf{0.7988} & \textbf{0.2213} & \textbf{0.9373} %\textbf{0.8026} 
    & \textbf{2.5571} & \textbf{0.3213} & \textbf{0.8207}
    \\ 

    \bottomrule[1.2pt]
  \end{tabular}
  }
  % \vspace{-8pt}
  
\end{table}

\noindent\textbf{Number of Blending Offsets.}
We vary the number of blending offsets $T$ in the dynamic Gaussian blending basis to investigate its effect on animation. As shown in Tab.~\ref{table:ablation_blendingoffsets}, increasing the number of blending offsets improves expression transfer and reconstruction quality by providing more capacity to capture diverse expression-dependent appearance changes. However, it also reduces animation speed. Thus, the number of blending offsets balances animation quality and efficiency. We set $T=15$ in our experiments.

\begin{table}[tb]
\caption{Ablation study on the number of blending offsets in the dynamic Gaussian blending basis on VFHQ~\cite{xie2022vfhq}.
  }
  \label{table:ablation_blendingoffsets}

  \vspace{-6pt}
  \centering
  \resizebox{0.8\linewidth}{!}{
  \begin{tabular}{l|ccc|c} 
  \toprule[1.2pt]
  {\#Offsets}  
             & PSNR $\uparrow$  & AKD $\downarrow$ & AED $\downarrow$ & Animation FPS $\uparrow$
             \\
    \midrule
    5 
    & 20.33 & 2.5910 & 0.3295 & \textbf{135.58} % 158.08
    \\
    10 
    & 20.39 & 2.5774 & 0.3246 & 122.44 % 142.76
    \\
    
    % \midrule
    15 (Ours) 
    & \textbf{20.41} & \textbf{2.5571} & \textbf{0.3213} & 114.47 % 133.47
    \\ 

    \bottomrule[1.2pt]
  \end{tabular}
  }
  % \vspace{-8pt}
  
\end{table}

% \noindent\textbf{Additional ablations for pixel-aligned Gaussians.}

\begin{figure}[tb]
    \centering
  \includegraphics[width=1\linewidth]{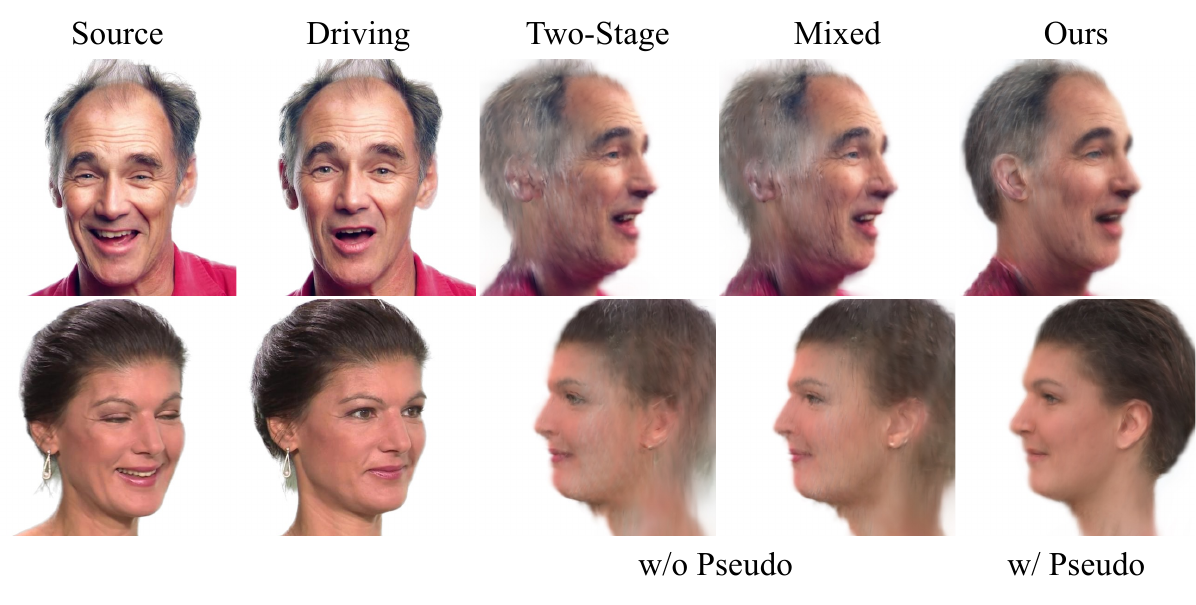}

  \vspace{-9pt}
  \caption{Qualitative ablation study on pseudo-multi-view supervision from 3D GAN inversion during the first training stage. We show animated novel-view results on VFHQ~\cite{xie2022vfhq}. Without this supervision, novel-view quality remains poor under both two-stage training and one-stage training with mixed monocular and multi-view data.
  % no matter using a two-stage training strategy or a one-stage training with mixed monocular and multi-view data, the novel view results remain poor.
  }
  % \vspace{-4pt}
  \label{fig:abla_3dganinvpng_supp}
\end{figure}

\noindent\textbf{Pseudo-Multi-View Supervision from 3D GAN Inversion.}
We evaluate the effect of pseudo-multi-view supervision from the pretrained 3D GAN inversion method~\cite{an2023panohead, bilecen2024dual} during the first training stage by removing this supervision. The results are shown in Fig.~\ref{fig:abla_3dganinvpng_supp}. Without pseudo-multi-view supervision in the first stage (column 3), the novel view results are so poor that they cannot be effectively refined even after incorporating real multi-view video data in the second stage. Because of the domain gap between real monocular data and multi-view data, novel-view renderings also remain poor when both types of data are used together in a single training stage (column 4), which is consistent with the observations reported in LCA~\cite{li2026large}. In contrast, pseudo-multi-view supervision enables the model to generate head avatars with coarse but complete 3D geometry, which can be effectively refined during the second training stage using real multi-view data.
%compare with using only monocular video at the first stage + normal second stage; using monocular video + real multi-view video in one stage

\begin{figure*}[tb]
    \centering
  \includegraphics[width=0.99\linewidth]{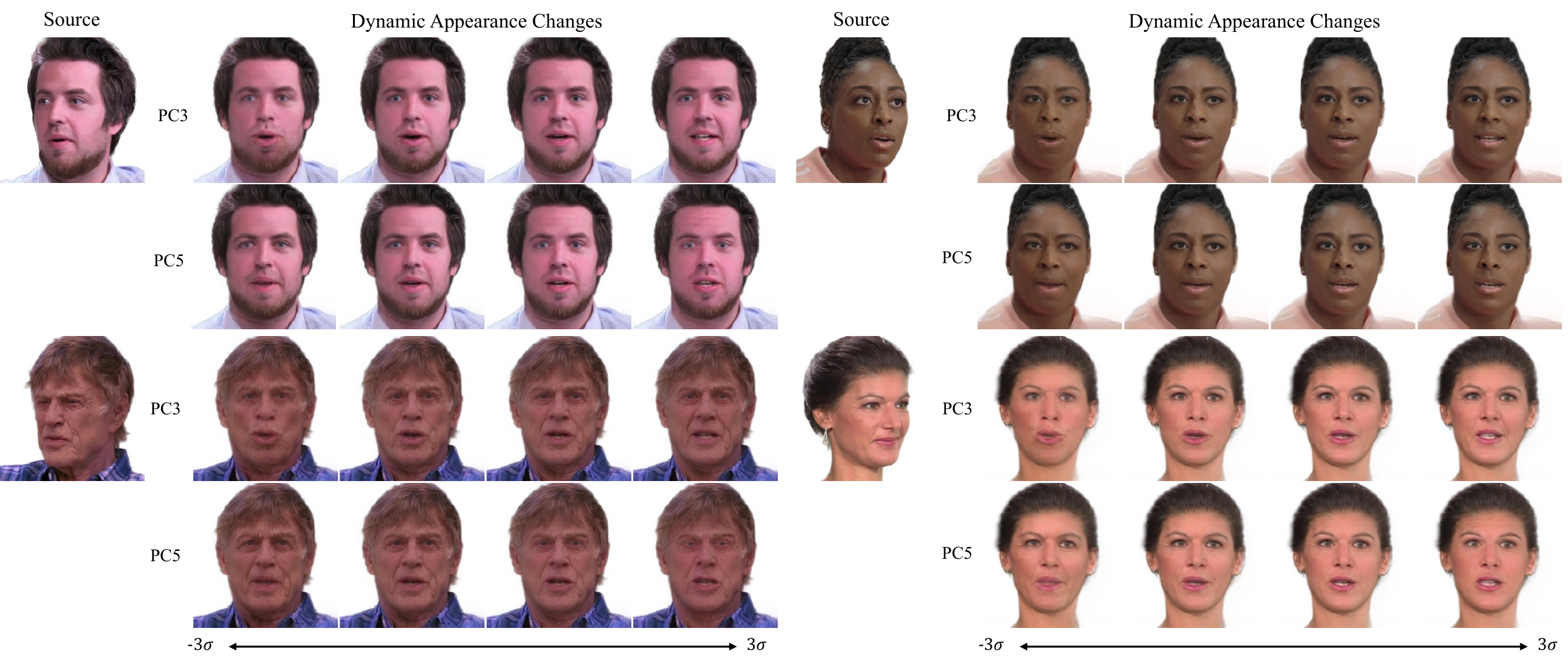}

  \vspace{-9pt}
  \caption{Additional visualizations of the dynamic Gaussian blending bases. We provide the rendered avatars with varying Gaussian colors and opacities by adjusting the weights of two visually distinct principal components (PC3 and PC5) of the blending offsets from $-3\sigma$ to $3\sigma$, where $\sigma$ denotes the standard deviation of the PC scores. By modifying only these weights to blend the bases while keeping all other Gaussian attributes fixed, the rendered avatars exhibit diverse expressions.
  }
  % \vspace{-4pt}
  \label{fig:vis_blend_pca_supp}
\end{figure*}

\noindent\textbf{Additional Visualization on Dynamic Gaussian Blending Bases.}
To visualize the generated dynamic Gaussian blending bases, we apply principal component analysis (PCA) to the blending offsets and extract the principal components (PCs), which capture the dominant modes of variation in the blending basis. Following~\cite{harkonen2020ganspace}, we vary the weight of one selected PC at a time. This allows us to modify the blending weights for the Gaussian blending bases and visualize the resulting appearance changes in the rendered avatars. As shown in Fig.~\ref{fig:vis_blend_pca_supp}, the rendered avatars exhibit diverse expressions despite changes only in Gaussian colors and opacities, while all other Gaussian attributes, including positions, remain fixed. These results highlight the importance of dynamic appearance modeling for producing natural expression changes and demonstrate the expressiveness of the generated dynamic Gaussian blending bases in capturing rich dynamic appearance details.

%We provide additional visualization results on the generated dynamic Gaussian blending bases in Fig.~\ref{fig:vis_blend_pca_supp}. 

% \noindent\textbf{Effect of the addition.} the selection of additional driving motion and camera pose to use for one-step inference

\begin{figure}[tb]
    \centering
  \includegraphics[width=1\linewidth]{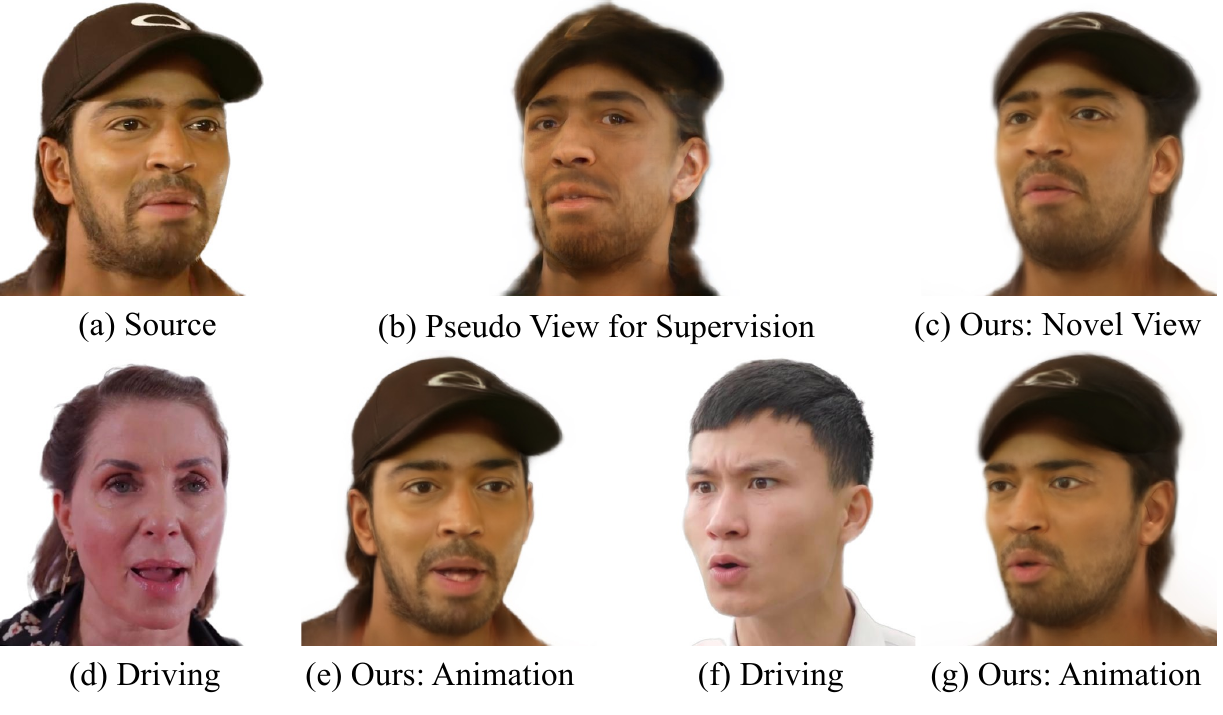}

  \vspace{-9pt}
  \caption{Failure cases of our method. Due to limitations in the pseudo-view supervision provided by 3D GAN inversion, our method fails to accurately reconstruct hat geometry. It also struggles to reproduce tongue movements and extreme expressions.
  }
  % \vspace{-4pt}
  \label{fig:failure}
\end{figure}

\section{Limitations}
\label{sec:limitations}
Although our method efficiently models dynamic appearance details for one-shot animatable head avatar reconstruction, it still has several limitations, with representative failure cases shown in Fig.~\ref{fig:failure}. For reconstruction, as shown in Fig.~\ref{fig:failure}b and c, our method is constrained by the pseudo-view supervision provided by the 3D GAN inversion method~\cite{an2023panohead, bilecen2024dual} and fails to accurately reconstruct hat geometry. Although we incorporate real multi-view videos during the second training stage to refine the 3D geometry, these videos rarely contain subjects wearing hats and therefore cannot effectively correct the reconstruction errors. For animation, our method relies heavily on the FLAME model, making its performance dependent on both the accuracy of FLAME parameter estimation and the model's ability to represent facial motions. For instance, it cannot model tongue movements, as shown in Fig.~\ref{fig:failure}d and e. Furthermore, the limited number of blending offsets in the generated blending bases and the limited representational capacity of the FLAME expression code hinder our method from reproducing large, extreme expression-dependent appearance changes, such as the transition from a nearly neutral expression to a deep frown, as shown in Fig.~\ref{fig:failure}f and~g. 
% To address these limitations, future works may consider using higher-quality pseudo-view supervision, exploring more expressive motion representations, and 

% 1. Cannot model appearance details of very large motions or remove it from the source image.
% 2. based on the FLAME model, limited by the tracking accuracy and FLAME expressiveness. It cannot model tongue motion.
% 3. Cannot model 3D hat shape.

\end{document}